\newif\ifmeta
\metatrue   
\ifdefined\buildcvpr\metafalse\fi  

\PassOptionsToPackage{table}{xcolor}
\ifmeta
  \documentclass{fairmeta}
\else
  \documentclass[10pt,twocolumn,letterpaper]{article}
  \usepackage[review]{cvpr}      
  \def\confName{CVPR}
  \def\confYear{2027}
\fi

\ifmeta
\usepackage[utf8]{inputenc}
\usepackage{url}
\usepackage{amsfonts}
\usepackage{amsmath}
\usepackage{nicefrac}
\usepackage{float}
\usepackage{wrapfig}
\usepackage{pifont}
\usepackage{enumitem}
\usepackage{array}
\usepackage{titletoc}
\newcolumntype{H}{>{\setbox0=\hbox\bgroup}c<{\egroup}@{}}  
\else
\usepackage[utf8]{inputenc}
\usepackage[T1]{fontenc}
\usepackage{amsfonts}
\usepackage{nicefrac}
\usepackage{microtype}
\usepackage{float}
\usepackage{cuted}  
\usepackage{multirow}
\usepackage{pifont}
\usepackage{array}
\usepackage{titletoc}
\newcolumntype{H}{>{\setbox0=\hbox\bgroup}c<{\egroup}@{}}  
\NewDocumentEnvironment{wrapfigure}{o m m}{\begin{figure}[t]\renewcommand{\vspace}[1]{}}{\end{figure}}
\NewDocumentEnvironment{wraptable}{o m m}{\begin{table}[t]\renewcommand{\vspace}[1]{}}{\end{table}}
\definecolor{cvprblue}{rgb}{0.21,0.49,0.74}
\AtBeginDocument{\crefname{appendix}{Supp.~Sec.}{Supp.~Secs.}\Crefname{appendix}{Supp.~Sec.}{Supp.~Secs.}%
  \crefname{subappendix}{Supp.~Sec.}{Supp.~Secs.}\Crefname{subappendix}{Supp.~Sec.}{Supp.~Secs.}%
  \crefname{subsubappendix}{Supp.~Sec.}{Supp.~Secs.}\Crefname{subsubappendix}{Supp.~Sec.}{Supp.~Secs.}}
\usepackage[pagebackref,breaklinks,colorlinks,allcolors=cvprblue]{hyperref}
\fi

\ifmeta
  \newenvironment{widefigure}[1][t]{\begin{figure}[#1]}{\end{figure}}
  \newenvironment{widetable}[1][t]{\begin{table}[#1]}{\end{table}}
\else
  \newenvironment{widefigure}[1][t]{\begin{figure*}[#1]}{\end{figure*}}
  \newenvironment{widetable}[1][t]{\begin{table*}[#1]}{\end{table*}}
\fi

\def\mypar#1{\vspace{0mm}\noindent\textbf{#1.}\hspace{0mm}} 

\definecolor{darkgreen}{RGB}{0,140,0}

\newcommand{\abstracttext}{\looseness=-1As image generation models scale to ever higher resolutions, global coherence, local detail, and texture fidelity become critical axes for generation quality. However, standard flow matching treats all spatial frequencies uniformly, ignoring the natural frequency hierarchy where high-frequency bands become indistinguishable from pure noise far earlier than coarse structures. We introduce \textbf{WaiT}, a \textbf{W}avelet-\textbf{a}ware \textbf{i}mage \textbf{T}ransformer that decomposes generation into coarse and fine bands via lossless wavelets. True to its name, the high-frequency bands \emph{``wait for the signal''}: staying pure noise until coarse structure has emerged, then joining the flow for joint refinement. Since standard FID discards fine-grained detail through aggressive downsampling, we introduce a more stringent three-axis evaluation protocol to assess quality at native resolution. On ImageNet 512$\times$512, WaiT achieves a pixel-space FID of 1.43 and is Pareto-optimal across all three axes, reducing sampling compute up to 50\%.
 With our largest  2B model, we set a new state-of-the-art FID of 1.3 for pixel-space models on ImageNet 512 resolution.
 Our formulation outperforms even the strongest latent-space models on texture fidelity, and scales seamlessly to high-resolution OpenImages and to video generation, achieving a state-of-the-art FVD of 0.84 on Kinetics-600 with no algorithmic modifications.}

\ifmeta
\title{WaiT for the Signal: Simple Frequency-Aware Flow-Matching}
\author[1,2]{Krunoslav Lehman Pavasovic}
\author[1,3]{Théophane Vallaeys}
\author[2]{Stéphane Mallat}
\author[2]{Giulio Biroli}
\author[1]{Luke Zettlemoyer}
\author[1]{Brian Karrer}
\author[1]{Jakob Verbeek}
\affiliation[1]{FAIR, Meta}
\affiliation[2]{École Normale Supérieure, Paris}
\affiliation[3]{Sorbonne University, Paris}
\abstract{\abstracttext}
\correspondence{Krunoslav Lehman Pavasovic at \email{krunolp@meta.com}.}
\metadata[Code]{\url{https://github.com/facebookresearch/WaiT}}
\else
\title{WaiT for the Signal: Simple Frequency-Aware Flow-Matching}
\author{Krunoslav Lehman Pavasovic$^{1,2}$ \quad Th\'eophane Vallaeys$^{1,3}$ \quad St\'ephane Mallat$^{2}$ \quad Giulio Biroli$^{2}$\\
Luke Zettlemoyer$^{1}$ \quad Brian Karrer$^{1}$ \quad Jakob Verbeek$^{1}$\\
$^{1}$FAIR, Meta \quad $^{2}$\'Ecole Normale Sup\'erieure, Paris \quad $^{3}$Sorbonne University, Paris}
\fi

\begin{document}

\maketitle

\newcommand{\teasercaptiontext}{
  \textbf{Evaluating WaiT  on ImageNet 512 and OpenImages 1024.} \textbf{Left:} Pareto fronts of compute vs.\   FID, 5-crop FID, and high-frequency FWD; WaiT  (green) dominates baselines at 2$\times$ lower compute. \textbf{Right:} WaiT generates sharper textures/finer details than both JiT and RAE.
  }

\ifmeta
\begin{figure*}[h]
  \centering
  \vspace{0mm}
  \includegraphics[width=.94\linewidth]{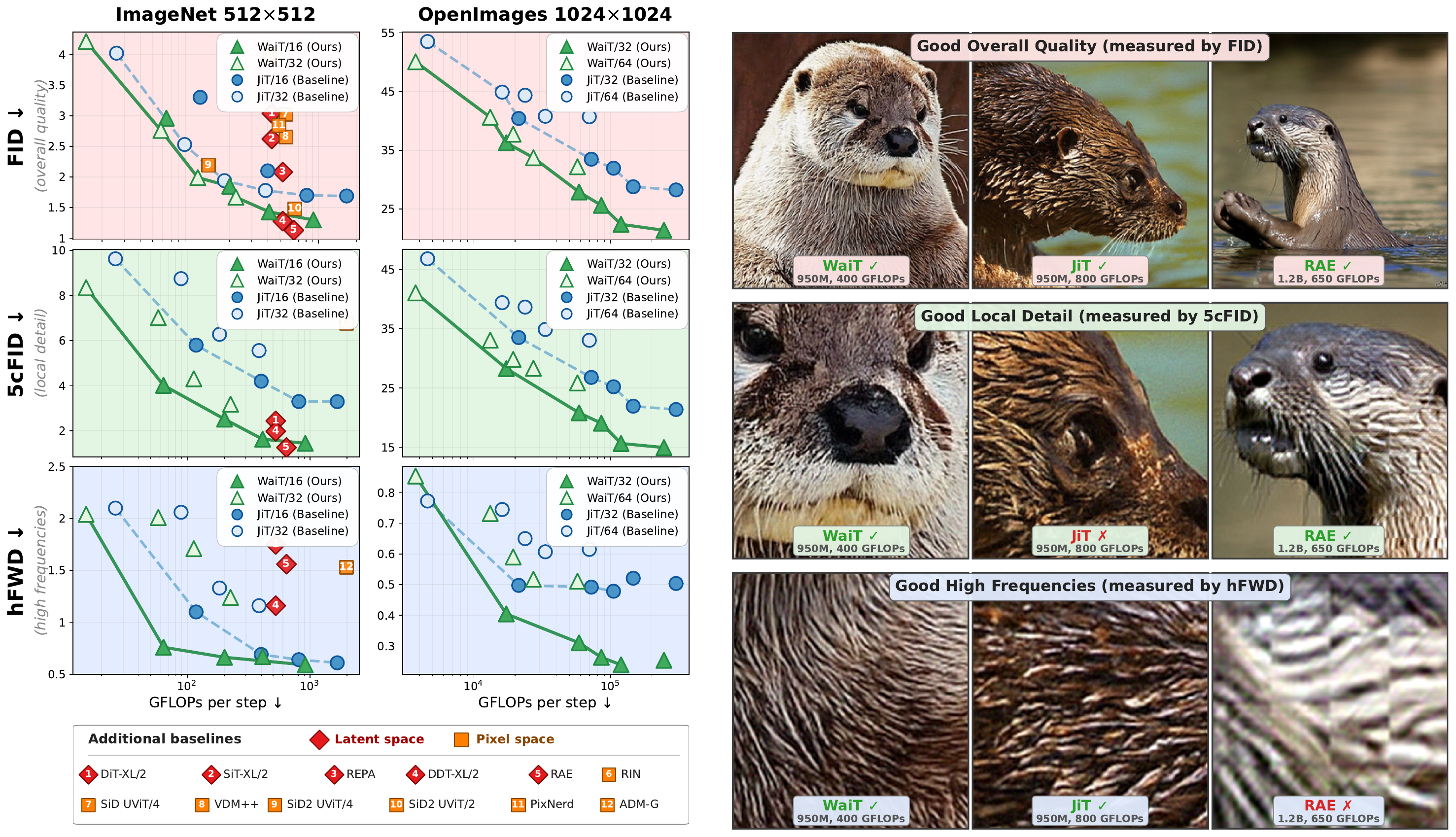}
  \vspace{-2mm}
  \caption{\teasercaptiontext}
  \label{fig:teaser}
  \vspace{-4.5mm}
\end{figure*}
\else
\begin{strip}
  \centering
  \includegraphics[width=\linewidth]{plots/pareto_with_qualitative.pdf}
  \captionsetup{hypcap=false}
  \captionof{figure}{\teasercaptiontext}
  \captionsetup{hypcap=true}
  \label{fig:teaser}
\end{strip}

\begin{abstract}
\abstracttext
\end{abstract}
\fi

\begin{widefigure}[!t]
    \centering
    \includegraphics[width=0.8\linewidth]{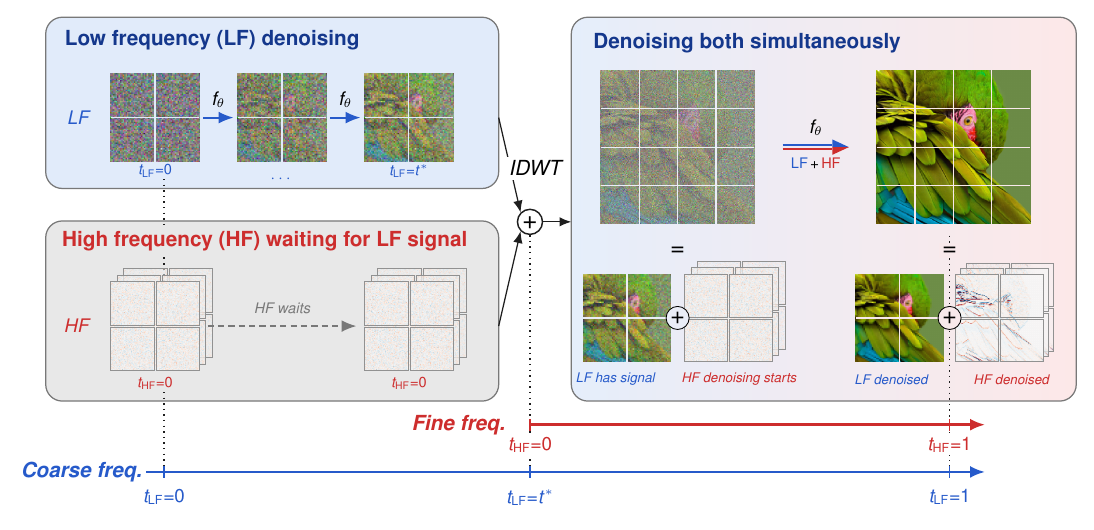}
    \caption{\textbf{Overview of WaiT   inference process.}
    The low-frequency (LF) wavelet band is denoised first along a coarse time $t_{\mathrm{LF}}$ to prioritize coarse structural coherence.
    The high-frequency (HF) bands wait as pure noise during this stage. At the crossover $t_{\mathrm{LF}}{=}t^{*}$, the partially denoised LF band and the noisy HF bands are combined via the inverse DWT and jointly denoised 
    yielding the final image.
    }
    \label{fig:method-overview}
    \ifmeta\vspace{-.5cm}\fi
\end{widefigure}

\section{Introduction}
\label{sec:intro}

\begin{wrapfigure}[24]{r}{0.38\textwidth}
  \vspace{-5mm}
  \centering
  \ifmeta\includegraphics[width=0.35\textwidth]{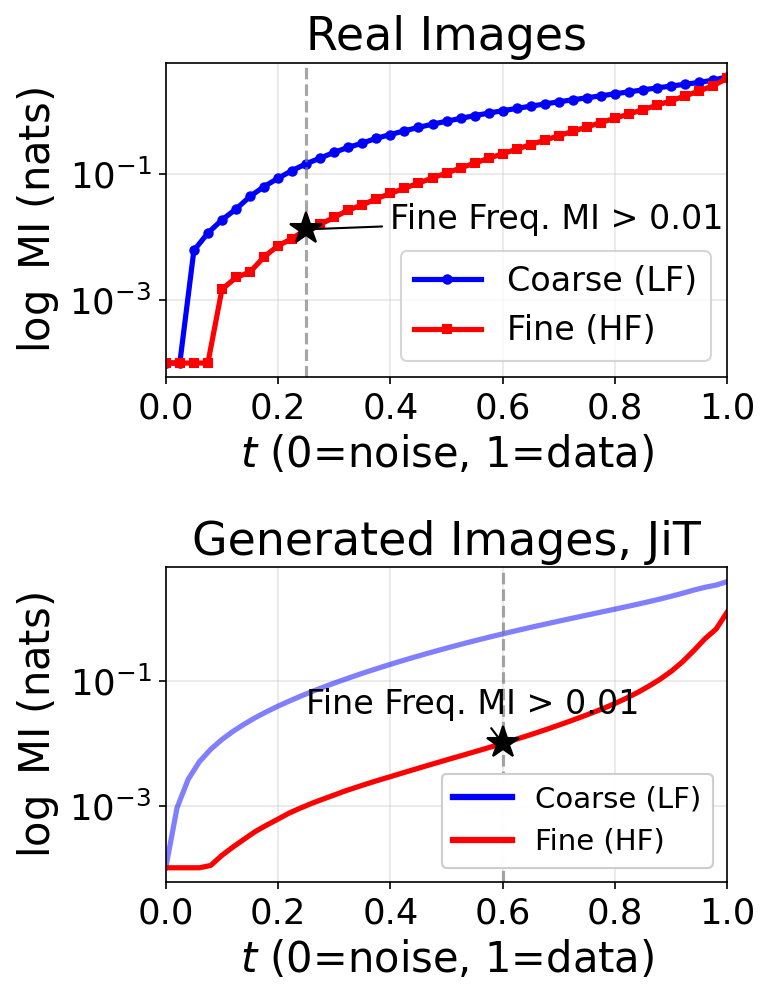}\else
  \includegraphics[width=0.49\linewidth]{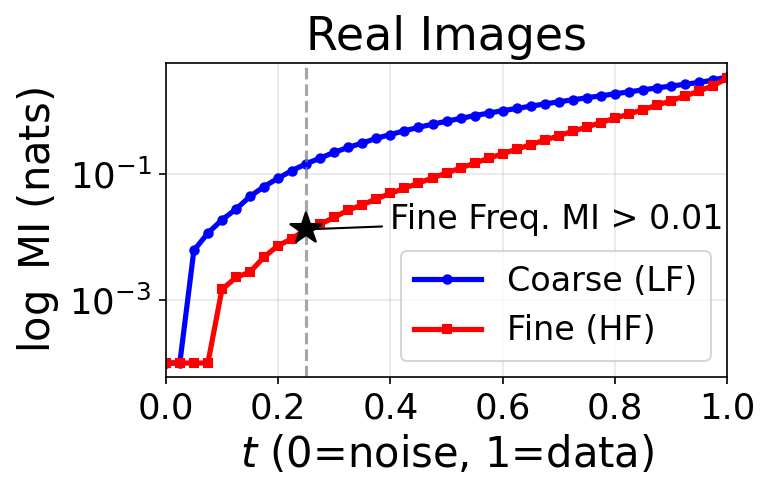}\hfill
  \includegraphics[width=0.49\linewidth]{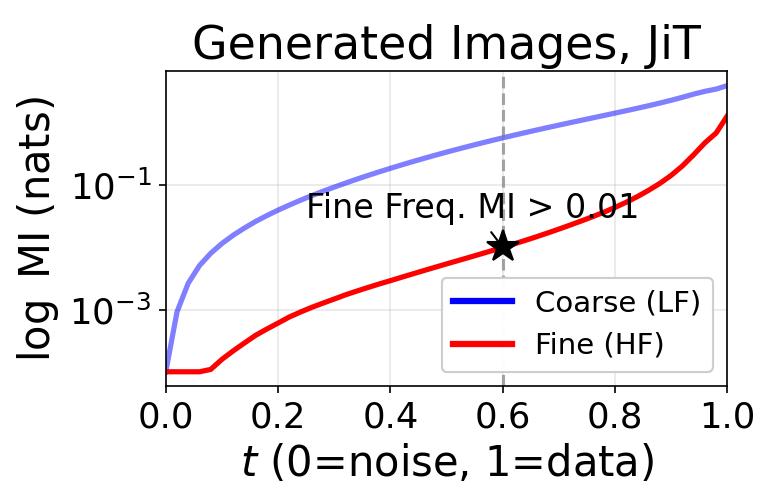}\fi
  \caption{\textbf{Mutual information (MI)} between noisy and clean wavelet bands. \textbf{\ifmeta Top\else Left\fi:} Forward process on real images. Fine (HF) bands lose MI far earlier than coarse (LF). The star marks $t^*{\approx}0.25$, where fine MI drops below 0.01 nats. \textbf{\ifmeta Bottom\else Right\fi:} The same asymmetry persists in backward process for JiT. See \ifmeta Appendix \else Supp.~Sec.~\fi\ref{app:mi_estimation} for details on MI estimation.
  }
  \label{fig:mi_comparison}
\end{wrapfigure}

Standard flow matching treats all spatial frequencies uniformly, but the natural power-law frequency hierarchy of images calls for a more nuanced approach because high-frequency content is destroyed by noise much earlier than coarse structure~\citep{falck2025fourierspaceperspective,ren2025xar,dieleman2024spectral}. We find that existing models spend a significant portion of their sampling trajectory on HF bands indistinguishable from noise (\Cref{fig:mi_comparison}), wasting modeling capacity precisely when global coherence is being established. We therefore propose to \textit{``wait for the signal''}, based on a long-standing principle from progressive compression standards like JPEG~2000~\citep{jpeg2000},
leveraging Discrete Wavelet Transform (DWT) to embed frequency awareness directly into the generative process.

Rather than engineering complex architectural pyramids, we simply decouple the noise schedules of coarse and fine bands in a pixel  diffusion model. By delaying the high-frequency generation, much like progressive compression delays high-frequency bit allocation \citep{shapiro1993embedded}, we allow the model to focus on coarse structural coherence without
wasting its modeling capacity on high-frequency components early in the denoising process.
We achieve this by defining two linear noise schedules: one over the full $[0,1]$ interval for the coarse band, and one over a limited  interval $[t^*,1]$ for the fine band.
The denoising process for inference naturally denoises low frequencies first, after which we inject high frequency noise at time $t^*$, combine both bands using the inverse wavelet transform, and then denoise both bands jointly.
This simple approach substantially improves generation quality while naturally unlocking significant computational savings across model scales, resolutions, and modalities.
See   \Cref{fig:method-overview} for an overview of our approach.

To measure performance for high-resolution image generation, we propose a new three-axis evaluation protocol.
As FID  downsamples  images to 299$\times$299, it inherently discards high-frequency textures and local details that distinguish true high-fidelity outputs, which is fundamentally insufficient as generation scales to 512$\times$512, 1024$\times$1024, and beyond.
To properly assess high-frequency generated content, we employ a three-axis evaluation protocol: in addition to FID, we use  5-crop FID (5cFID) for local structural detail via native-resolution crops, and high-frequency Fréchet Wavelet Distance (hFWD) to isolate pure texture fidelity.
See \Cref{app:metrics} for details on both metrics.

We implement our approach on the recent pixel-space JiT model~\citep{Li2025}.
Through extensive experiments, we show that this approach yields excellent results:
an FID of 1.43, 5cFID of 1.63, and hFWD of 0.67 on ImageNet 512$\times$512, while reducing sampling GFLOPs by up to 50\% compared to the JiT baseline.
Notably, on pure texture fidelity (hFWD) we outperform both state-of-the-art pixel- and latent-space methods.
With our largest 2B model we set a new best  FID  of 1.3 for ImageNet 512 resolution for pixel-space models.
Similar trends are observed for class-conditional generation on OpenImages (\Cref{fig:teaser}).
Beyond class-conditional generation, WaiT transfers to pixel-space text-to-image synthesis at 1024$\times$1024, matching or exceeding JiT on almost all metrics at up to 3$\times$ higher throughput. The same recipe carries over unchanged to the temporal domain, where spatiotemporal wavelets reach a state-of-the-art FVD of 0.84 on Kinetics-600.

Our main contributions are as follows:
(1) We introduce Wavelet-aware image Transformer, a pure pixel-space approach that embeds the natural frequency structure of images into the denoising process: a principled inductive bias that improves generation quality while significantly reducing computational cost.
(2) We employ a three-axis evaluation protocol (FID, 5cFID, hFWD) capturing global coherence, local detail, and texture fidelity, revealing complementary weaknesses in both pixel- and latent-space methods that standard FID cannot detect. The protocol is validated, showcasing strong correlation with human preference judgments (\Cref{app:perceptual}).
(3) We demonstrate Pareto optimality among pixel-space models across all three metrics on ImageNet 512 and OpenImages 512 and 1024 resolution.
(4) We show that the framework extends directly to text-to-image and video generation. On text-to-image at 1024$\times$1024, WaiT improves over the JiT baseline across almost all metrics while achieving up to 3$\times$ higher throughput; on video, spatiotemporal wavelets yield a state-of-the-art FVD of 0.84 on Kinetics-600 using the same core recipe.

\section{Related Work}
\label{sec:related}

\textbf{Pixel-Space and Cascaded Diffusion.} Pure pixel-space modeling has recently re-emerged as competitive with latent approaches, led by Simpler Diffusion (SiD2)~\citep{Hoogeboom2024} and Just image Transformers (JiT)~\citep{Li2025}. Beyond diffusion, JetFormer~\citep{tschannen2025jetformer} models raw images and text jointly with a normalizing-flow--based autoregressive prior, and is also trained end-to-end without a pretrained autoencoder. To scale to high resolutions, cascaded models such as CDM~\citep{ho22cascaded}, PixelFlow~\citep{chen2025pixelflow}, and Pyramidal Flow Matching~\citep{jin2024pyramidalflowmatchingefficient} generate in stages, but suffer a train--test mismatch:
high-frequencies carry faint signal at training time yet are replaced by noise at inference.
Existing fixes (noise augmentation, upsamplers) are ad-hoc; our frequency-domain construction removes this mismatch in a principled manner.

\mypar{Architectural Frequency Awareness} Several methods exploit frequency separation via specialized architectural designs: DeCo~\citep{Ma2025} pairs a DiT~\citep{peebles23iccv} for LF semantics with a pixel decoder for details, PixelDiT~\citep{yu2025pixeldit} and DiP~\cite{chen2025dip} use dual pathways, Edify Image~\citep{Atzmon2024} processes Laplacian-pyramid scales in cascade, and LapFlow~\cite{Zhao2026} in parallel. In contrast, WaiT requires no architectural changes beyond adding a resolution embedding to the existing time and class conditioning, solving frequency awareness purely through the noise schedule on a standard transformer architecture design.

\mypar{Frequency-Domain Representations} DCTdiff~\citep{Ning2024} models diffusion in DCT space, but relies on lossy HF truncation and a global, uniform noise schedule. Unlike DCTdiff, our approach is lossless, using band-specific schedules to exploit the temporal asymmetry of noise corruption. Closer to us, FreqFlow~\citep{ren2025xar} also observes that flow matching generates low frequencies before high frequencies, but injects this prior via a two-branch architecture with band-specific losses and frequencies-derived features fed into a pixel-level branch. In contrast, our approach acts purely on the noise schedule and requires no architectural changes.

\section{The WaiT Approach}
\label{sec:method}

\looseness=-1Our design rests on two well-established observations. First, the natural frequency hierarchy of images~\citep{falck2025fourierspaceperspective,dieleman2024spectral} implies that high-frequency signals are overwhelmed by noise far sooner than coarse structures, which standard flow matching ignores. We further confirm this asymmetry empirically in trained pixel-space models (\Cref{fig:mi_comparison}).
Second, scale-wise factorization is theoretically shown to be more stable than joint modeling~\citep{Guth2022}, suggesting fine details should follow rather than lead the coarse structure during generation. We therefore formulate the generative process in the wavelet domain to explicitly control the noise schedule at different frequencies and \emph{delay} the fine bands.
In practice, we implement this approach on top of the transformer architecture of JiT \citep{Li2025}.

\subsection{Frequency-specific Noise Schedules}
\label{sec:schedules}

We utilize the Discrete Wavelet Transform (DWT) to separate the signal into distinct bands. Given an image $x \in \mathbb{R}^{D}$, a single-level DWT yields a low-frequency (LF) approximation band $x_{LF} \in \mathbb{R}^{D/4}$ and high-frequency (HF) detail bands $x_{HF} \in \mathbb{R}^{3D/4}$. Because the Haar DWT is orthogonal and invertible, it serves as a lossless bridge between coarse structure and fine detail without requiring external models. In our framework, the LF band assumes the role of the spatially compressed space used by latent generative models; unlike learned autoencoders, the wavelet transform is lossless, invertible, and requires no training.
In this paper we use the Haar wavelet\footnote{For readers less familiar with wavelets: the LF band of a single-level 2D Haar transform is, up to scale, $2{\times}2$ average pooling, and the three HF bands store the horizontal, vertical, and diagonal differences needed to invert the transform losslessly.} for simplicity.
In \Cref{app:wavelet}, we find WaiT robust to the choice of wavelet family.

Standard flow matching models~\citep{lipman22arxiv} define a uniform interpolation $z_t = t \cdot x + (1-t) \cdot \epsilon$ for $t \in [0,1]$. We decouple this process into band-specific signal mixing coefficients $t_{LF}$ and $t_{HF}$. We identify $t_{LF}$ with the global time $t$ (so the LF band follows the standard schedule), and let the HF band lag behind via the delayed schedule introduced below; we therefore use $t$ and $t_{LF}$ interchangeably from here on.
The initial step is to decompose the clean image $x \in \mathbb{R}^{D}$ into its frequency components: $(x_{LF}, x_{HF}) = \text{DWT2}(x)$. Before flow matching we normalize the LF band by a scalar $S_{LF}$, writing $\tilde{x}_{LF} = x_{LF}/S_{LF}$, while the HF bands are left unnormalized (see \textit{Band Normalization} below). We then apply band-specific interpolation to these components:

$$ t_{HF} = \max\left(0, \frac{t_{LF} - t^*}{1 - t^*}\right). $$

Both bands then follow a straightforward linear interpolation using their respective timescales:\ifmeta$$ z_{LF, t} = t_{LF} \cdot \tilde{x}_{LF} + (1-t_{LF}) \cdot \epsilon_{LF}, \qquad
 z_{HF, t} = t_{HF} \cdot x_{HF} + (1-t_{HF}) \cdot \epsilon_{HF}.$$\else
\begin{gather*} z_{LF, t} = t_{LF} \cdot \tilde{x}_{LF} + (1-t_{LF}) \cdot \epsilon_{LF}, \\
 z_{HF, t} = t_{HF} \cdot x_{HF} + (1-t_{HF}) \cdot \epsilon_{HF}.\end{gather*}\fi
\looseness=-1In this formulation, $\tilde{x}_{LF}$ is the normalized LF band and $x_{HF}$ the HF wavelet coefficients of the clean image $x$, and $\epsilon_{LF} \sim \mathcal{N}(0, \sigma_{LF}^2 I)$, $\epsilon_{HF} \sim \mathcal{N}(0, I)$ are independent Gaussian noise samples drawn directly in the wavelet domain, with $z_{LF,t}$ and $z_{HF,t}$ having $D/4$ and $3D/4$ entries, respectively. Here $\sigma_{LF}$ is the resolution-dependent noise scale of JiT~\citep{Li2025} applied at the LF resolution $\text{res}/2$: $\sigma_{LF} = \text{res}/512$ for OpenImages and text-to-image, and $\sigma_{LF} = 1.5 \cdot \text{res}/512$ for ImageNet (\Cref{app:noise_scale}).
For $t_{LF} \leq t^*$, we have $t_{HF} = 0$, so the HF band is pure unit-variance noise: $z_{HF,t} = \epsilon_{HF}$. During sampling, we inject fresh noise $\epsilon_{HF} \sim \mathcal{N}(0, I)$ at $t^*$, which matches this distribution exactly.

This mathematical alignment ensures the sampling distribution perfectly matches the training distribution at the transition point $t^*$, resolving the train-test discrepancy of discontinuous schedules (see \Cref{fig:schedules}),
resulting in a structured, colored Gaussian noise in the pixel space.

\mypar{Band Normalization}
\label{sec:normalization}
The DWT redistributes signal energy unevenly: the LF band produces values in $[\text{-}2, 2]$ (for input normalized to $[\text{-}1, 1]$), while HF bands are naturally small-valued ($\sigma{\sim}0.05$--$0.19$ on ImageNet 256). As mentioned, we normalize the LF band before flow matching:
$$ \tilde{x}_{LF} = x_{LF} \,/\, S_{LF},$$
where $S_{LF}$ is the 95th percentile of absolute LF coefficients over the training set (e.g., $S_{LF} = 1.94$ for ImageNet 256 resolution).
We prefer $p_{95}$ over the maximum (outlier-sensitive) or standard deviation (heavy-tailed). Note that centering is not needed: inputs are normalized to $[-1,1]$ and the linear Haar low-pass filter preserves the mean, so the LF band only needs rescaling by $S_{LF}$. Furthermore, HF bands are left unnormalized: a few large coefficients already match the noise schedule, while the near-zero bulk reflects natural sparsity--preserving it keeps the model focused on structural edges rather than amplifying numerical noise. See App.~\ref{app:normalization} for further discussion, where we also ablate several normalization schemes.

\subsection{Training Objective and Modifications}
\label{sec:training}

\mypar{Training objective}
We adopt the JiT~\citep{Li2025} training objective: $x$-prediction with $v$-loss, applied separately to each phase. Using the relation $v_\theta = (x_\theta - z)/(1-t)$, the $v$-loss becomes a reweighted $x$-loss. In our case, supervision targets the wavelet bands of $x$ rather than $x$ itself, so each frequency can be weighted by its own noise level, something a single pixel-space loss against $x$ cannot express.

\textit{Coarse objective}: The model operates on the normalized LF band, predicting normalized $\tilde{x}_{LF}$ from the noisy input $z_{LF,t} = t_{LF} \cdot \tilde{x}_{LF} + (1-t_{LF}) \cdot \epsilon_{LF}$ where $t_{LF} = t$:
$$ \mathcal{L}_{\text{coarse}} = \mathbb{E}_{t, x, \epsilon} \left[ (1-t_{LF})^{-2}\, \mathrm{MSE}\big(x_\theta(z_{LF,t}, t), \tilde{x}_{LF}\big) \right], $$
where $\mathrm{MSE}$ denotes the mean squared error over the coefficients of the band.
We train the coarse phase over the full range $t\in[0,1]$, however during sampling coarse is only until $t<t^{*}$.
We find this to lead to  important improvements, and ablate this choice in our experiments (\Cref{fig:designing_wait}(c), \Cref{tab:interval_app}).
Working directly with the LL band keeps the two phases in nearly the same domain: the LL band of a wavelet transform is a low-pass, locally averaged version of the image, so it stays spatially coherent and image-like, and after zero-mean-preserving filtering and the global LF normalization its mean and scale closely match those of pixels. Together with joint training of both phases and a resolution embedding, this allows a single backbone to work seamlessly with both the coarse LF domain and the full-resolution pixel domain without a noticeable domain shift.

\textit{Fine objective}: The model operates in full-resolution pixel space for $t > t^*$, with $t = t^* + (1-t^*)\,u$ and $u$ drawn from the same logit-normal distribution as in the coarse phase. The noisy input $\tilde{z}_t = \text{IDWT}(z_{LF,t}\cdot S_{LF}, z_{HF,t})$  has \emph{frequency-dependent noise levels}: the LF band is interpolated at $t_{LF} = t$, while the HF band uses its accelerated schedule $t_{HF} = (t_{LF} - t^*)/(1 - t^*)$.
A single forward pass $\hat{x} = x_\theta(\tilde{z}_t, t)$ produces a pixel-space prediction, which we decompose via DWT into $(\hat{x}_{LF}, \hat{x}_{HF}) = \text{DWT2}(\hat{x})$, normalize the LF prediction as $\hat{\tilde{x}}_{LF} := \hat{x}_{LF}/S_{LF}$, and supervise with band-specific reweighting:
\ifmeta
$$ \mathcal{L}_{\text{fine}} = \mathbb{E}_{t > t^*, x, \epsilon} \left[ (1-t_{LF})^{-2}\, \mathrm{MSE}\big(\hat{\tilde{x}}_{LF}, \tilde{x}_{LF}\big) + (1-t_{HF})^{-2}\, \mathrm{MSE}\big(\hat{x}_{HF}, x_{HF}\big)\right], $$
\else
\begin{equation*}\begin{aligned} \mathcal{L}_{\text{fine}} = \mathbb{E}_{t > t^*, x, \epsilon} \Big[ &(1-t_{LF})^{-2}\, \mathrm{MSE}\big(\hat{\tilde{x}}_{LF}, \tilde{x}_{LF}\big) \\ &+ (1-t_{HF})^{-2}\, \mathrm{MSE}\big(\hat{x}_{HF}, x_{HF}\big)\Big], \end{aligned}\end{equation*}
\fi
where again $\mathrm{MSE}$ is taken over band coefficients (each band contributes on the same scale regardless of its size).
Here $\tilde{x}_{LF}$ is the target normalized LF band and $\hat{\tilde{x}}_{LF}$ its prediction, while $x_{HF}$ is the target HF band and $\hat{x}_{HF}$ its prediction.
The DWT applied to the network output is computationally negligible compared to the forward pass.

\mypar{Resolution conditioning} Similar to PixelFlow~\citep{chen2025pixelflow}, we encode the side length of the current token grid (e.g., 8 and 16 for a /32 model at $512{\times}512$) via a standard sinusoidal embedding and MLP, adding it to the time and class conditioning so the model can distinguish denoising phases.

\subsection{Sampling}
\label{sec:sampling}

\begin{wraptable}{r}{0.5\textwidth}
  \vspace{-3.2em}
  \centering
  \footnotesize
  \setlength{\tabcolsep}{3pt}
  \caption{WaiT aligns the generative process with image hierarchy: low frequencies lead the structural synthesis, while high frequencies wait for the signal--evolving on a separate timescale.}
  \label{tab:teaser_pseudocode}
  \scalebox{1}{%
  \begin{tabular}{@{}rl@{}}
  \toprule
  \multicolumn{2}{l}{\textbf{Wavelet-aware image Transformer (WaiT)}} \\
  \midrule
  1. & $z_\text{low},\, z_\text{high} \sim \mathcal{N}(0, I)$ \\[3pt]
  2. & \textbf{for} $t = 0 \to t^*$: \hfill \textit{\color{gray} Coarse flows; Fine waits} \\
     & \quad $z_\text{low} \leftarrow \text{Denoise}(z_\text{low},\, t)$ \\[3pt]
  3. & \textbf{for} $t = t^* \to 1$: \hfill \textit{\color{gray} Fine emerges; Joint refinement} \\
     & \quad $[z_\text{low},\, z_\text{high}] \leftarrow \text{Denoise}(z_\text{low},\, z_\text{high},\, t)$ \\[3pt]
  4. & \textbf{return} IDWT($z_\text{low}$, $z_\text{high}$) \\
  \bottomrule
  \end{tabular}}

\end{wraptable}
\textbf{Two-phase generation.}
At inference, generation proceeds in two phases (\Cref{tab:teaser_pseudocode}): (1)~ODE integration for $0\leq t<t^*$ at low resolution (LF band only), then (2)~un-normalize the LF band, inject fresh unit-variance HF noise, map to full resolution via IDWT, and continue the ODE for $t^*<t\leq 1$, as shown in Figure~\ref{fig:schedules} as Delayed Linear.
Phase~0 operates at $4\times$ fewer tokens (for a single-level DWT), reducing the compute cost of sampling.

\mypar{Step allocation} Given a total budget of $K$ ODE steps, we spend a fraction $\rho$ of them in Phase~0: $K_1 = \mathrm{round}(\rho K)$ uniformly spaced steps on $[0, t^*]$, and the remaining $K - K_1$ uniformly spaced steps on $[t^*, 1]$. Since Phase~0 steps operate at $4\times$ fewer tokens, larger $\rho$ directly reduces total GFLOPs. See \Cref{app:step_allocation} for the values used in our experiments.

\section{Experiments}
\label{sec:experiments}

\textbf{Datasets.} For ablations and main results we consider class-conditional generation based on the common ImageNet-1k~\citep{russakovsky2015imagenet} at  256 and 512 resolution setup.
Going beyond 512 resolution on ImageNet is  not meaningful as  only 5.3\% of the images have a shortest side $\geq$512.
So the vast majority of reference images would be upsampled, injecting interpolation artifacts that corrupt precisely the high-frequency statistics we are interested in.
 We therefore curate natively 512 and 1024 resolution  subsets from OpenImages V6~\citep{kuznetsova2020openimages} with similar number of classes and images as ImageNet-1k. The dataset construction details required to reproduce the exact train/val splits, are provided in \ifmeta Appendix~\else Supp.~Sec.~\fi\ref{sec:dataset_construction}.

\textbf{Metrics.} To measure generation quality we report three different metrics.
Standard \textbf{FID}  quantifies \textbf{global coherence}, as its  downsampling to 299$\times$299 discards high-frequency information. 
We  introduce \textbf{5-crop FID} to measure \textbf{local detail} by calculating the FID with statistics computed over five distinct, 299 native-resolution crops (corners + center) instead of downsampling. We also tested random native-resolution crops and obtained almost identical results; we adopt the fixed corners+center scheme for determinism and reproducibility. 
Complementing this, we use \emph{high-frequency FWD} (\textbf{hFWD}), a dedicated metric for \textbf{texture fidelity}, which computes the
Fr\'{e}chet Wavelet Distance~\citep{veeramacheneni2025fwd} while explicitly excluding the low frequencies (DC wavelet packet). This exclusion is critical because the LF component contains no high-frequency signal and would otherwise dominate the metric, making it redundant with FID.
These three metrics (global coherence, local detail, and texture fidelity) together provide a necessary and comprehensive evaluation of modern generative models for high-fidelity, high-resolution synthesis.
We verify in \Cref{app:perceptual} that these metrics  align with human preference using  the PIPAL  benchmark~\citep{gu2020pipal}.


\begin{widefigure}[t]
  \centering
  \begin{subfigure}[b]{0.285\textwidth}
    \centering
    \fcolorbox{blue!40}{white}{\includegraphics[width=0.98\linewidth]{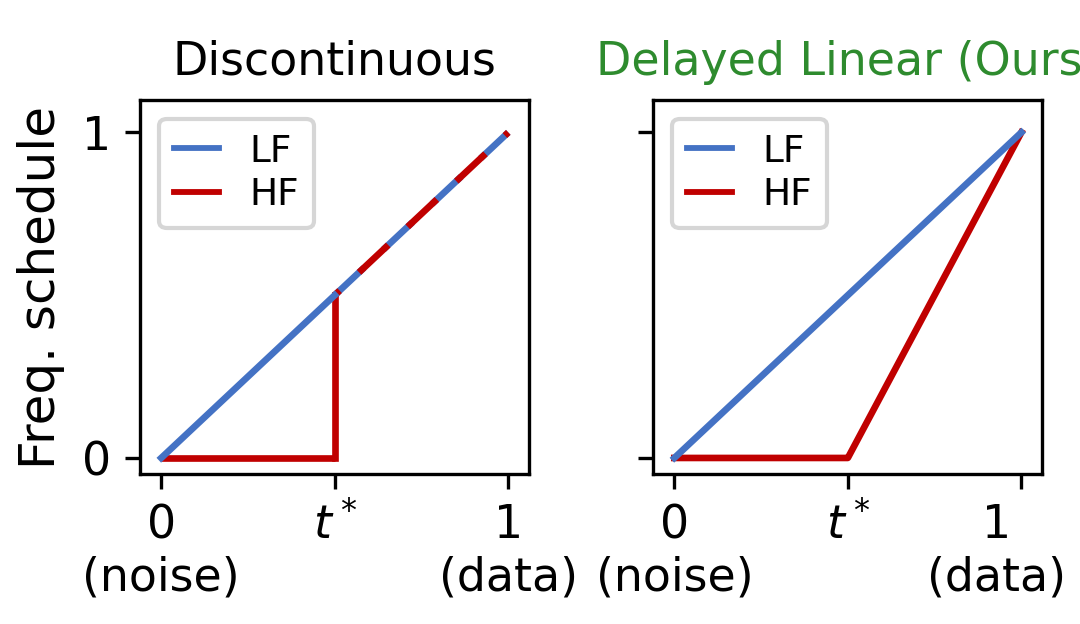}}
    \caption{\textcolor{blue!70!black}{Schedule choice.}}
    \label{fig:schedules}
  \end{subfigure}\hfill%
  \begin{subfigure}[b]{0.285\textwidth}
    \centering
    \fcolorbox{orange!60}{white}{\includegraphics[width=0.98\linewidth]{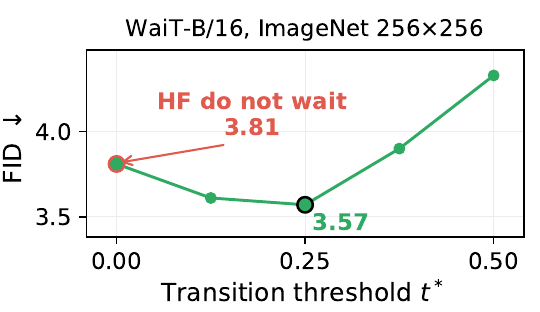}}
    \caption{\textcolor{orange!80!black}{Optimal $t^*$.}}
    \label{fig:tstar_sensitivity}
  \end{subfigure}\hfill%
  \begin{subfigure}[b]{0.40\textwidth}
    \raisebox{0.45in}{%
    \begin{minipage}{\linewidth}
    \centering
    \scriptsize
    \setlength{\tabcolsep}{2.5pt}
    \begin{tabular}{@{}clcc@{}}
      \toprule
      & Design choice & FID\,$\downarrow$ & $\Delta$ \\
      \midrule
      0 & Na\"ive two-stage JiT & 7.51 & --- \\
      \rowcolor{blue!10}
      1 & + Delayed linear sched.~\textbf{(a)} & 5.43 & $-2.08$ \\
      2 & + Global coarse $[0,1]$ & 4.33 & $-1.10$ \\
      \rowcolor{orange!18}
      3 & + Optimal $t^*\!=\!0.25$~\textbf{(b)} & \textbf{3.57} & $-0.76$ \\
      \bottomrule
    \end{tabular}
    \end{minipage}}
    \caption{Cumulative ablation.}
    \label{tab:ablation}
  \end{subfigure}
  \caption{\textbf{Designing WaiT.} Two most important design choices are (a) the multi-res. \emph{schedule}, where the delayed linear schedule lets HF bands smoothly emerge from $t^*$ while finishing jointly at $t=1$, and (b) the choice of threshold $t^*$, optimal at $t^*{=}0.25$ where HF mutual information vanishes (\Cref{fig:mi_comparison}); at $t^*{=}0$ the method reduces to standard JiT. (c) Cumulative effect of these choices plus global coarse training on ImageNet 256$\times$256 with WaiT-B/16; the final model halves the FID while saving 40\% GFLOPs.}
  \label{fig:designing_wait}
\end{widefigure}

\subsection{Design ablation}
\label{sec:ablation}

\textbf{How to denoise in the frequency domain?} A natural starting point is a discontinuous two-stage cascade as in PixelFlow~\citep{chen2025pixelflow} and Pyramidal Flow~\citep{jin2024pyramidalflowmatchingefficient}: a coarse stage on $[0,t^*]$ followed by  HF noise injection at $t^*$ (leftmost panel of \Cref{fig:schedules}). This naïve two-stage baseline (\Cref{fig:designing_wait}(c), row~0) uses a fixed $t^*{=}0.5$ and trains the coarse stage only on $[0,t^*]$; rows~1--3 then add our improvements cumulatively.
However, the hard handoff creates a train--test mismatch at $t^*$ (FID $7.51$). Replacing it with a \textbf{delayed linear} schedule which couples LF and HF \emph{continuously} in a single jointly trained model cuts FID to $5.43$ ($-2.08$); see Fig.\ref{fig:schedules} and Fig.\ref{tab:ablation}. Further schedule variants (cascaded, mixed) are compared in \Cref{tab:schedules_app}.

\textbf{What is the best range to train the LF band?}
We  find  \emph{global coarse training} over the full  $[0,1]$  interval to improve over using  $[0,t^*]$ as in~\citep{chen2025pixelflow,jin2024pyramidalflowmatchingefficient} ($-1.10$ FID; row~2 of Fig.\ref{tab:ablation}). This is a simple but intuitive choice that exposes the model to LF tokens across the full trajectory, letting it learn how HF depends on the underlying coarse structure rather than treating LF as an isolated subproblem.

\begin{table}[t]
  \centering
  \ifmeta\small\else\footnotesize\setlength{\tabcolsep}{2pt}\fi
  \renewcommand{\arraystretch}{0.9}
  \caption{Results on class-cond. ImageNet 512x512.
  $\dagger$\, using SSL or text-supervised latent space.
  Results taken from original papers, except for models marked with $*$ which we trained ourselves using open-source code.
  5cFID\,=\,5-crop FID. hFWD\,=\,high.\ freq.\ FWD. We evaluate the latter metrics using open-source weights, or mark '-' where not available.
  }\ifmeta\vspace{-1mm}\fi
  \label{tab:pixel_comparison}
  \begin{tabular*}{\linewidth}{@{\extracolsep{\fill}}lccc@{\hspace{2\tabcolsep}}Hcc@{}}
    \toprule
    \textbf{Method} & \textbf{params} & \textbf{GFLOPs} & \textbf{FID$\downarrow$} & \textbf{IS$\uparrow$} & \textbf{5cFID$\downarrow$} & \textbf{hFWD$\downarrow$} \\
    \midrule
    \multicolumn{7}{l}{\textit{Latent-space models}} \\
    DiT-XL/2 \cite{peebles23iccv} & 675+49M & 525 & 3.04 & 240.8 & 2.44 & 1.75 \\
    SiT-XL/2 \cite{ma2024sit}& 675+49M & 525 & 2.62 & 252.2 & - & - \\
    REPA, SiT-XL/2$^{\dagger}$ \cite{yu2025repa} & 675+49M & 525 & 2.08 & 274.6 & - & - \\
    FreqFlow \cite{ren2025xar} & 507+49M & N/A & 2.02 & - & - & - \\
    DDT-XL/2$^{\dagger}$ \cite{wang2025ddt}& 675+49M & 525 & 1.28 & \bf 305.1 & 1.98 & \bf 1.16 \\
    RAE, DiT$^{\text{DH}}$-XL/2$^{\dagger}$ \cite{zheng202rae} & 839+415M & 642 & \textbf{1.13} & 259.6 & \bf 1.26 & 1.56 \\
    \midrule
    \multicolumn{7}{l}{\textit{Pixel-space models}} \\
    ADM-G \cite{dhariwal21nips}& 559M & 1983 & 7.72 & 172.7 & 6.77 & 1.53 \\
    RIN \cite{jabri2023rin} & 320M & 415 & 3.95 & 216.0 & - & - \\
    SiD, UViT/4 \cite{hoogeboom2023simplediffusion}& 2B & 555 & 3.02 & 248.7 & - & - \\
    PixNerd, XL/16$^{\dagger}$ \cite{wang2025pixnerd}& 700M & 583 & 2.84 & 245.6 & - & - \\
    VDM++, UViT/4 \cite{kingma2023vdm++}& 2B & 555 & 2.65 & \bf 278.1 & - & - \\
    DiP-XL/32 \cite{chen2025dip}& 631M & N/A & 2.31 & - & - & - \\
    SiD2, UViT/4 \cite{Hoogeboom2024}& N/A & 137 & 2.19 & - & - & - \\
    Pixel DiT \cite{yu2025pixeldit}& 797M & 1352 & 1.81 & - & 2.19 & 2.19 \\
    SiD2, UViT/2 & N/A & 653 & \textbf{1.48} & - & - & - \\
    \midrule
    \textbf{JiT-B/32} \cite{Li2025} & 133M & 26 & 4.02 & 271.0 & 9.63 & 2.10 \\
    \textbf{JiT-B/16}$^*$ & 133M & 118 & 3.30 & 266.3 & 5.80 & 1.10 \\
    \textbf{JiT-L/32} & 462M & 89 & 2.53 & 299.9 & 8.75 & 2.06 \\
    \textbf{JiT-L/16}$^*$ & 462M & 400 & 2.10 & 292.7 & 4.20 & 0.69 \\
    \textbf{JiT-H/32} & 956M & 183 & 1.94 & \bf 309.1 & 6.28 & 1.33 \\
    \textbf{JiT-H/16}$^*$ & 956M & 810 & 1.70 & 307.5 & 3.30 & 0.64 \\
    \textbf{JiT-G/32} & 2B & 384 & 1.78 & 306.8 & 5.56 & 1.16 \\
    \textbf{JiT-G/16}$^*$ & 2B & 1665 & \bf 1.69 & 301.5 & 3.30 & \bf 0.61 \\
    \midrule
    \multicolumn{7}{l}{\textit{Wavelet-aware image Transformer (ours)}} \\
    \textbf{WaiT-B/32} & 133M & 15 & 4.21 & 261.8 & 8.35 & 2.04 \\
    \textbf{WaiT-B/16} & 133M & 64 & 2.96 & 273.3 & 4.02 & 0.76 \\
    \textbf{WaiT-L/32} & 462M & 58 & 2.76 & 296.5 & 7.02 & 2.01 \\
    \textbf{WaiT-L/16} & 462M & 201 & 1.85 & TBD & 2.52 & 0.66 \\
    \textbf{WaiT-H/32} & 956M & 113 & 1.99 & 300.3 & 4.31 & 1.71 \\
    \textbf{WaiT-H/16} & 956M & 411 & 1.43 & 277.5 & 1.63 & 0.67 \\
    \textbf{WaiT-G/32} & 2B & 225 & 1.67 & \bf 302.5 & 3.18 & 1.24 \\
    \textbf{WaiT-G/16} & 2B & 914 & \textbf{1.30} & 300.1 & \textbf{1.45} & \textbf{0.59} \\
    \bottomrule
  \end{tabular*}
\end{table}

\textbf{Does delaying HF help, and by how much?} Our mutual-information analysis (\Cref{fig:mi_comparison}) shows that HF carries no learnable signal at small $t$, so  $t^*>0$ should preserve quality while cutting compute, and free up modeling capacity to improve LF denoising in the first stage. Sweeping $t^*$ confirms this (\Cref{fig:tstar_sensitivity}): $t^*\!=\!0$ is equivalent to a standard JiT, while $t^*\!=\!0.25$ gives a further $-0.24$ FID. 

\subsection{Main results: class-conditional generation}
\label{sec:hf_quality}

\mypar{ImageNet 512$\times$512}
We compare WaiT  with state-of-the-art generative image models in \Cref{tab:pixel_comparison}.
WaiT improves over all previous pixel-space models at matched compute, as well as most latent-space baselines, and matches JiT FID at roughly half the inference compute. Notably, WaiT-H/16 reaches FID 1.43 at 411 GFLOPs, and WaiT-G/16 sets a new pixel-space SOTA of 1.30 FID. \Cref{fig:teaser} shows that  WaiT  consistently improves the Pareto-front over JiT across all three metrics; see Fig.\ref{fig:model_comparison} for qualitative comparison with JiT and RAE, and \Cref{fig:bandlimited_generations,fig:lighthouse_comparison,fig:starfish_full_appendix} in the \ifmeta appendix\else supplementary material\fi\ for additional samples.

\begin{widefigure}[t]
  \centering
  \includegraphics[width=0.9\linewidth]{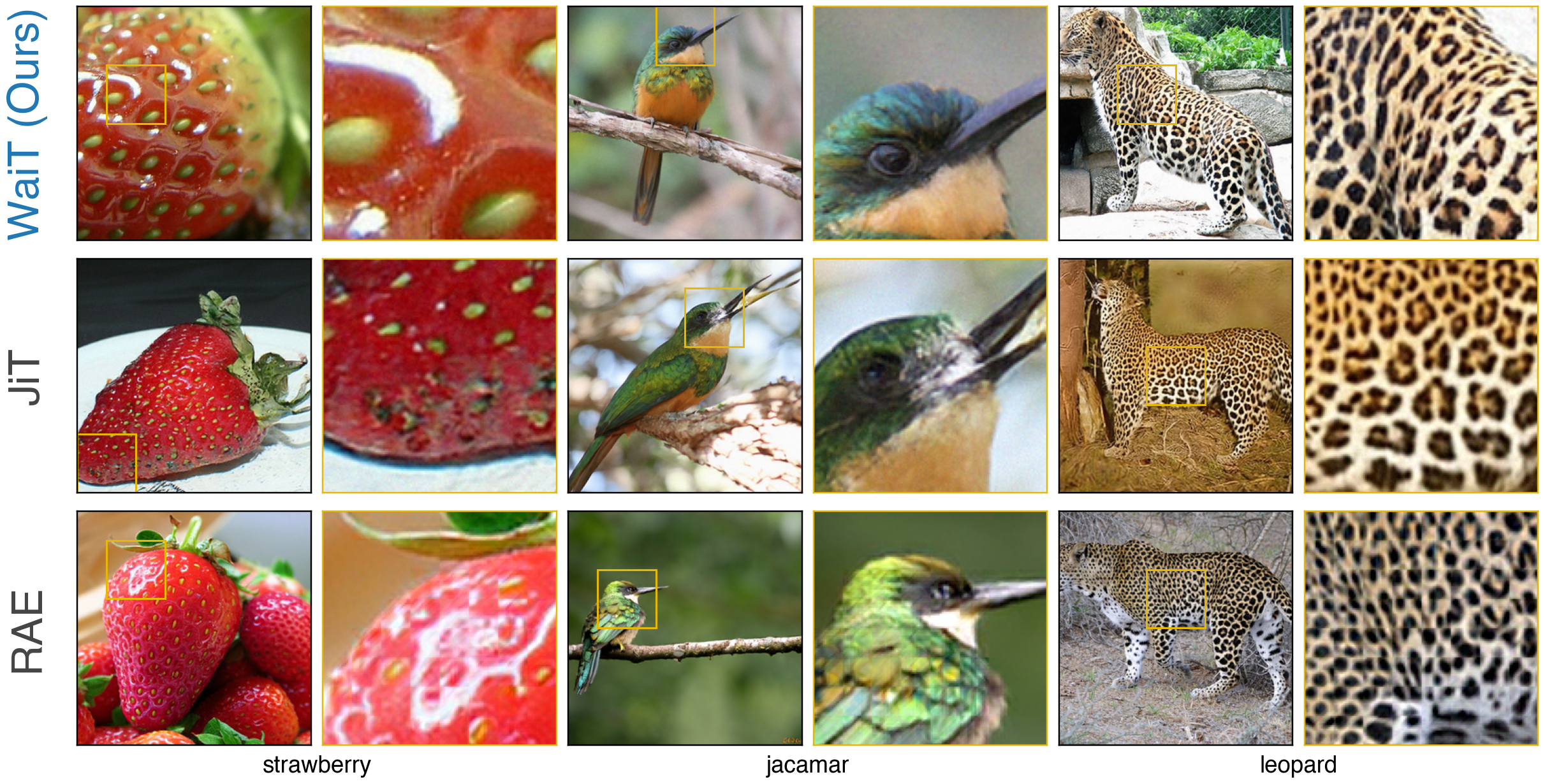}
  \caption{\textbf{Qualitative comparison across state-of-the-art methods} on ImageNet 512$\times$512.
  WaiT-H/16, JiT-H/16 and RAE on three classes (strawberry, jacamar, leopard); each class displays the full resolution image with a yellow rectangle marking the zoomed crop.
  The crops highlight a consistent pattern: \textbf{JiT} visibly struggles with fine local detail (blurred strawberry seeds, jacamar eye and beak, and leopard's facial detail), while \textbf{RAE} produces sharp textures but distorts high-frequency \emph{structure} (with visible checkerboard artifacts). \textbf{WaiT}  performs well on both axes, preserving fine detail \emph{and} the underlying high-frequency structure.
  }
  \label{fig:model_comparison}
  \ifmeta\vspace{-3mm}\fi
\end{widefigure}

\mypar{OpenImages-1M}
\looseness=-1\Cref{fig:teaser} reports compute vs.\ image quality on OpenImages 1024 for WaiT and JiT from B/64 to G/32. WaiT dominates JiT's Pareto front by a large margin, matching its (5-crop) FID at $\sim\!2\times$ lower compute; while JiT's hFWD plateaus around 0.5 already at B/32, WaiT keeps benefiting from more inference compute and pushes hFWD well below 0.3.
The \ifmeta appendix\else supplementary material\fi\ provides analogous 512-resolution plots with similar trends (\Cref{fig:pareto_openimages}), Haar coefficient distributions (\Cref{fig:wavelet_hf}), and qualitative comparisons at 512 and 1024 (\Cref{app:openimages_pareto}).

\newcommand{\tiitablebody}{  \centering
  \scriptsize
  \setlength{\tabcolsep}{3pt}
  \caption{\textbf{Comparison of WaiT with JiT on text-to-image generation.}
  We consider two data preparation pipelines, and train H/32 models at 1024 resolution. }
  \label{tab:t2i_results}
  \resizebox{\linewidth}{!}{%
  \begin{tabular}{@{}llccccccc@{}}
    \toprule
    \textbf{Encoders} & \textbf{Method} & \textbf{FID$\downarrow$} & \textbf{5cFID$\downarrow$} & \textbf{hFWD$\downarrow$} & \textbf{CLIP$\uparrow$} & \textbf{GenEval$\uparrow$} & \textbf{DPG$\uparrow$} & \textbf{imgs/s$\uparrow$} \\
    \midrule
    \multirow{2}{*}{SigLIP + Qwen3-VL + T5-v1.1-XXL}
      & JiT-H/32           & 5.35 & 8.97 & 3.47 & 26.71          & 0.442          & \textbf{0.824} & 0.10 \\
      & \textbf{WaiT-H/32} & \textbf{5.13} & \textbf{8.10} & \textbf{3.16} & \textbf{26.75} & \textbf{0.491} & 0.823          & \textbf{0.29} \\
    \midrule
    \multirow{2}{*}{MetaCLIP + Llama 3.2 Vision + Llama 3.1 }
      & JiT-H/32           & 5.20 & 5.50 & 2.33 & 39.10          & 0.419          & 0.798          & 0.12 \\
      & \textbf{WaiT-H/32} & \textbf{4.74} & \textbf{5.33} & \textbf{2.15} & \textbf{39.43} & \textbf{0.460} & \textbf{0.805} & \textbf{0.28} \\
    \bottomrule
  \end{tabular}}
}

\subsection{Further extensions of WaiT}
\label{sec:extensions}
{A useful side effect of only modifying the noise schedule is that WaiT is easy to combine with other changes, since the backbone stays untouched. We check this in two ways: adding a third frequency level, and applying the same recipe in a latent space. In both cases we reuse the WaiT recipe directly, running each setting once and without any hyperparameter tuning.

\mypar{Multi-level transitions} Our formulation extends directly from a single ($2$-level) transition to multiple levels. On OpenImages-1M at $1024\times1024$ (L/32), we apply a $3$-level schedule out of the box, normalizing both the first and second LF bands by their respective $p_{95}$ constants and setting the two transition points to $t^*=\{0.15, 0.3\}$ (\Cref{tab:multilevel}). The additional level further increases compute savings and improves both high-frequency metrics (5cFID and hFWD), at a marginal cost in FID.

\mypar{Transfer to latent space} Although WaiT is motivated by pixel space, the same recipe applies unchanged in a learned latent space. We apply it to DDT/XL-2~\citep{wang2025ddt} out of the box, training WaiT+DDT from scratch directly at $512\times512$ for only $500$k steps, whereas the DDT baseline is trained for $1.28$M steps at $256\times256$ and then finetuned for a further $500$k steps at $512\times512$ (\Cref{tab:latent}). Despite using less than a third of the training budget, WaiT+DDT matches the baseline FID ($1.33$ vs.\ $1.28$), improves 5cFID ($1.77$ vs.\ $1.98$), and reduces inference compute by ${\sim}35\%$ ($329$ vs.\ $525$ GFLOPs); matching the training budget or light tuning would likely improve this further.}

\begin{widetable}[t!]
\centering
\small
\ifmeta\else\begin{minipage}[t]{0.525\textwidth}\centering\scriptsize\setlength{\tabcolsep}{2pt}\fi
\caption{\textbf{Multi-level transitions} on OpenImages-1M $1024\times1024$ (L/32), applied out of the box, FID evaluated on 50k images.}
\label{tab:multilevel}
\begin{tabular}{@{}lcccc@{}}
\toprule
Configuration & FID ($\downarrow$) & 5cFID ($\downarrow$) & hFWD ($\downarrow$) & Compute savings ($\uparrow$) \\
\midrule
WaiT ($2$ levels, $t^*{=}0.25$)              & \textbf{9.74}  & 21.71          & 0.918          & 18.5\% \\
WaiT ($3$ levels, $t^*{=}\{0.15, 0.3\}$)     & 10.11          & \textbf{16.03} & \textbf{0.867} & \textbf{33.0\%} \\
\bottomrule
\end{tabular}
\ifmeta

\vspace{1.5em}

\else\end{minipage}\hfill\begin{minipage}[t]{0.465\textwidth}\centering\scriptsize\setlength{\tabcolsep}{2pt}\fi
\caption{\textbf{Out-of-the-box transfer to latent space} (DDT/XL-2) on ImageNet $512\times512$.}
\label{tab:latent}
\begin{tabular}{@{}lccccc@{}}
\toprule
Method & FID ($\downarrow$) & 5cFID ($\downarrow$) & hFWD ($\downarrow$) & GFLOPs ($\downarrow$) & Training steps \\
\midrule
DDT~\citep{wang2025ddt}  & \textbf{1.28} & 1.98          & \textbf{1.16} & 525          & 1.28M\,+\,500k \\
WaiT\,+\,DDT  & 1.33          & \textbf{1.77} & 1.20          & \textbf{329} & \textbf{500k (from scratch)} \\
\bottomrule
\end{tabular}
\ifmeta\else\end{minipage}\par\bigskip
\tiitablebody\fi
\end{widetable}

\newcommand{\tiiparta}[1]{#1{Pixel-space text-to-image generation}
\label{sec:t2i}
\label{app:t2i}
We now test the WaiT recipe on text-to-image generation at 1024 resolution. To do so, we curate a ${\sim}40$M-image dataset from fully public sources (SA-1B~\citep{kirillov2023sam}, DataComp Multimodal~\citep{gadre2024datacomp}, and OpenImages~\citep{kuznetsova2020openimages}), filtered by resolution, aesthetic and watermark scoring, recaptioned with VLMs; we use two fully public data preparation pipelines, based on SigLIP~\citep{zhai2023sigmoid}\,+\,Qwen3-VL~\citep{Qwen3VL2025}\,+\,T5-v1.1-XXL~\citep{raffel2020exploring} and on MetaCLIP~\citep{xu2023demystifying}\,+\,Llama 3.2 Vision~\citep{dubey2024llama3}\,+\,Llama 3.1~\citep{dubey2024llama3}, and give the complete construction recipe in \Cref{app:t2i_data}.}
\newcommand{\tiipartb}{We train JiT-H/32 and WaiT-H/32 models on this dataset mixture at the native 1024 resolution.
In \Cref{tab:t2i_results} we report results as measured over a held-out evaluation set, adding CLIPScore~\cite{hessel2021clipscore} to measure alignment between the prompt and the generated images, as well as GenEval and DPG metrics.
Using both data preparation pipelines, WaiT yields improved results than JiT, while achieving up to 3$\times$ higher throughput.
We present several qualitative examples comparing WaiT with JiT in \Cref{fig:t2i_combined} and \Cref{fig:t2i_appendix}, which can both be found in the \ifmeta appendix\else supplementary material\fi.

\ifmeta\begin{widetable}[t]
\tiitablebody\end{widetable}\fi}
\newcommand{\tiifigure}{\begin{widefigure}[t]
  \centering
  \makebox[\textwidth]{%
    \makebox[0.5\textwidth]{\footnotesize\textbf{SigLIP + Qwen3-VL + T5-v1.1-XXL}}%
    \makebox[0.5\textwidth]{\footnotesize\textbf{MetaCLIP + Llama 3.2 Vision + Llama 3.1}}%
  }\\[2pt]
  \includegraphics[width=\textwidth]{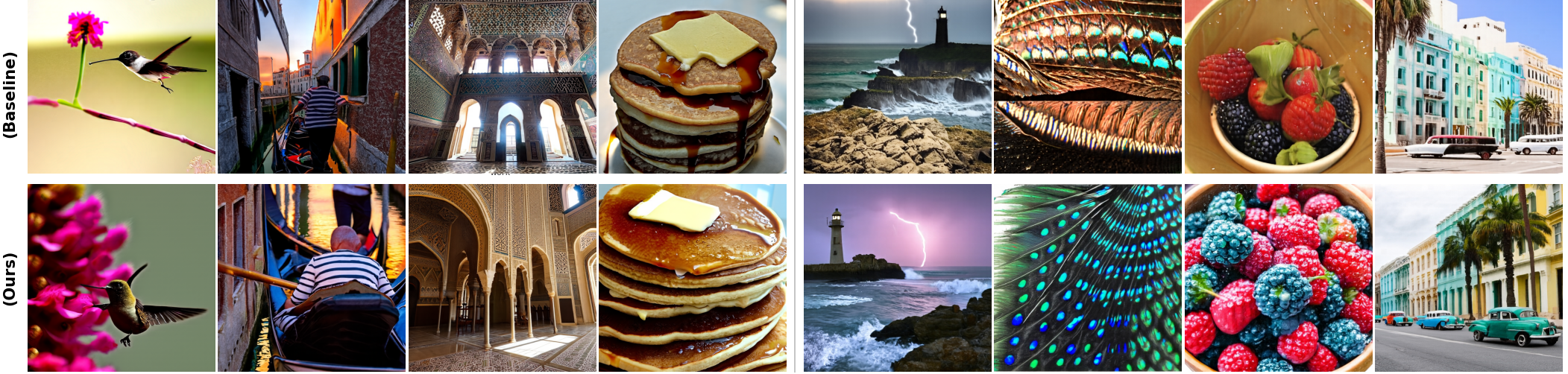}
  \caption{\textbf{Text-to-image samples at 1024 resolution.} Per-column prompts and per-row method/encoder details are provided in \Cref{app:t2i_captions}.}
  \label{fig:t2i_combined}
\end{widefigure}}

\tiiparta{\subsection}

\newcommand{\kineticstable}{  \caption{ \textbf{Kinetics-600} (128$\times$128, 5$\to$16 frames). FVD on full validation set.}
    \label{tab:video_kinetics}
  \centering
  \scriptsize
  \begin{tabular}{@{}lcc@{}}
    \toprule
    Model & FVD ($\downarrow$)& GFLOPs / $10^{2}$ \\
    \midrule
    RIN \cite{jabri2023rin}& 10.8 & 1,200 \\
    MAGVIT-v2 \cite{yu2024language} & 4.3 & 120 \\
    W.A.L.T. \cite{gupta2023walt} & 3.3 & 120 \\
    UL (S) \cite{heek26ul} & 1.7 & 300 \\
    UL (M) & 1.3 & 700 \\
    \midrule
    JiT-B/8 & 1.50 & 300 \\
    \textbf{WaiT-B/8} & \textbf{1.45} & \textbf{210} \\
    JiT-XL/8 & 0.89 & 1,580 \\
    \textbf{WaiT-XL/8} & \textbf{0.84} & \textbf{1,110} \\
    \bottomrule
  \end{tabular} }

\ifmeta\else\begin{wraptable}{r}{0.45\textwidth}
  \centering
  \caption{\textbf{Video results:} Taichi-HD (128$\times$128) FVD on 5K generated videos vs.\ training set, and Kinetics-600 (256$\times$256, 5$\to$16 frames) FVD on full validation set.}
  \label{tab:video_taichi}
  \label{tab:video_kinetics_256}
  \scriptsize
  \setlength{\tabcolsep}{2pt}
  \begin{tabular}{@{}l cc @{\hspace{3pt}} ccc @{\hspace{3pt}} cc@{}}
  \toprule
  & \multicolumn{5}{c}{\textbf{Taichi-HD}} & \multicolumn{2}{c}{\textbf{Kinetics-600}} \\
  \cmidrule(r){2-6}\cmidrule(l){7-8}
  & \multicolumn{2}{c}{Uncond. (B/8)} & \multicolumn{3}{c}{Cond. 1$\rightarrow$16 (B/8)} & \multicolumn{2}{c}{Cond. 5$\rightarrow$16 (XL/32)} \\
  \cmidrule(r){2-3}\cmidrule(lr){4-6}\cmidrule(l){7-8}
  & FVD ($\downarrow$) & GFLOPs & FVD ($\downarrow$) & LPIPS ($\downarrow$) & GFLOPs & FVD ($\downarrow$) & GFLOPs / $10^{2}$ \\
  \midrule
  JiT & 28.92 & 21,260 & 19.92 & 0.297 & 21,260 & 1.32 & 371 \\
  \textbf{WaiT} & \textbf{27.13} & \textbf{16,624} & \textbf{19.75} & \textbf{0.292} & \textbf{16,694} & \textbf{1.20} & \textbf{319} \\
  \bottomrule
  \end{tabular}
\end{wraptable}\fi

\tiipartb\ifmeta

\tiifigure\fi

\subsection{Video generation results}
\label{sec:video}

Finally, we show that WaiT directly generalizes to video generation: the wavelets now decomposes three axes (height, width, time), reducing the coarse-phase token count $8\times$.
We use two datasets for our experiments: Taichi-HD~\citep{siarohin2019taichi} with $\sim$3k videos of people performing Tai Chi, and  Kinetics-600~\citep{carreira2018kinetics600} with $\sim$500k videos. In all video experiments we use a temporal patch size of 4 frames, while the model size and spatial resolution vary by experiment (from B/8 at $128\times128$ up to XL/32 at $256\times256$).
To measure the generated video quality we use the FVD metric~\citep{unterthiner2019fvd}. In case of generation conditioned on one or more real frames, we also measure LPIPS~\citep{zhang18cvpr} metric comparing the frames of the generated continuation with those of the ground-truth one.

\ifmeta
\begin{table}[t]
  \scriptsize
  \setlength{\tabcolsep}{3pt}
  \begin{minipage}[t]{0.35\textwidth}
  \caption{\textbf{Taichi-HD} (128$\times$128) FVD on 5K generated videos vs.\ training set.
  }
  \label{tab:video_taichi}
  \centering
  \begin{tabular}{@{}lccc@{}}
  \toprule
  & FVD ($\downarrow$) & LPIPS ($\downarrow$) & GFLOPs \\
  \midrule
  \multicolumn{2}{l}{\textbf{Unconditional Generation}} & & \\ 
  JiT-B/8 & 28.92 & - & 21,260 \\
  \textbf{WaiT-B/8} & \textbf{27.13} & - & \textbf{16,624} \\
  \midrule
  \multicolumn{3}{l}{\textbf{Conditional Generation} (1$\rightarrow$16 frames)}  & \\ 
  JiT-B/8 & 19.92 & 0.297 & 21,260 \\
  \textbf{WaiT-B/8} & \textbf{19.75} & \textbf{0.292} & \textbf{16,694} \\
  \bottomrule
  \end{tabular}
  \end{minipage}\hfill
  \begin{minipage}[t]{0.33\textwidth}
\kineticstable
  \end{minipage}\hfill
  \begin{minipage}[t]{0.28\textwidth}
  \caption{\textbf{Kinetics-600} (256$\times$256, 5$\to$16 frames). FVD on full validation set.}
  \label{tab:video_kinetics_256}
  \centering
  \begin{tabular}{@{}lcc@{}}
    \toprule
    Model & FVD ($\downarrow$) & GFLOPs / $10^{2}$ \\
    \midrule
    JiT-XL/32 & 1.32 & 371 \\
    \textbf{WaiT-XL/32} & \textbf{1.20} & \textbf{319} \\
    \bottomrule
  \end{tabular}
  \end{minipage}
\end{table}
\fi

\textbf{Taichi-HD.}
We consider generation of 16 frames, either  unconditionally or conditioned on an additional initial frame, where the final frame is discarded for FVD evaluation.
In \Cref{tab:video_taichi} we report  results for WaiT and the  JiT baseline.
 WaiT improves over JiT in both settings, while leading to about 22\% compute reduction.
 Qualitative examples are shown in  Fig.~\ref{fig:taichi_samples} in the \ifmeta appendix\else supplementary material\fi.

\textbf{Kinetics-600.}
Here  we  condition on 5 frames and  generate the remaining 11 frames (16 in total). For FVD evaluation, we use the initial 5 frames with the subsequent 11 generated frames, following~\cite{gupta2023walt,heek26ul,jabri2023rin,yu2024language}.
In \Cref{tab:video_kinetics}\ifmeta\else{} (supplementary material)\fi\ we compare WaiT with JiT (we trained JiT models using open source code) as well as previous results from the literature.
While JiT already improves the FVD of 1.7 from  Unified Latents (small)~\citep{heek26ul} to 1.5 at equal compute, WaiT  further reduces the FVD to 1.45 while reducing compute by 30\%.
With the larger WaiT-XL/8 we are able to push the FVD further down to a new state-of-the-art result of 0.84\ifmeta\else\ (comparison to other models in \Cref{tab:video_kinetics} in the appendix.)\fi.
Uncurated samples  are shown in Figure~\ref{fig:kinetics_samples} in the \ifmeta appendix\else supplementary material\fi.
We additionally train XL/32 models on Kinetics-600 at $256\times256$ (\Cref{tab:video_kinetics_256}), where WaiT again improves quality while reducing compute.

On both benchmarks, WaiT matches or improves video quality while reducing compute by up to 30\%.
These results demonstrate that the core benefits of WaiT,
improved generation and reduced compute,
transfer seamlessly across modalities. 
In the video experiments we used the Phase~0 step fraction $\rho$ optimized for images, but are confident that minor, targeted tuning could yield further substantial GFLOP savings at minimal cost to video quality.

\section{Conclusion}
\label{sec:conclusion}

\looseness=-1We introduced the Wavelet-aware image Transformer (WaiT), a principled extension to pixel-space flow matching models. WaiT employs a delayed and accelerated noise schedule for the HF band, withholding its denoising until low-frequency signals are available. Governed by a single transition point $t^*$ and a lossless, parameter-free wavelet transform, WaiT integrates seamlessly into existing architectures like JiT.
Despite this simplicity, WaiT achieves Pareto-optimal compute--quality trade-offs across global structure (FID), local detail (5-crop FID), and texture fidelity (hFWD) on ImageNet 512 and OpenImages 512/1024. On ImageNet 512, we set a new state-of-the-art FID of 1.3 for pixel-space models, with WaiT-H/16 improving FID by ${\sim}30\%$ and 5-crop FID by ${\sim}60\%$ over the compute-matched JiT-L/16 (411 vs.\ 400 GFLOPs). We observe similar improvements in pixel-space text-to-image modeling at 1024 resolution. When applied to video, WaiT achieves state-of-the-art FVD on Kinetics-600 with up to 30\% compute reduction, demonstrating that a single, unified recipe suffices for high-fidelity image and video generation.

We hope our approach will serve as a foundation for future pixel-space generative models operating at even higher resolutions and across additional modalities.

\ifmeta
\section*{Acknowledgments}
This work has received funding from the French government, managed by the National Research Agency (ANR), under the France 2030 program with the reference ANR-23-IACL-0008. Furthermore, this paper is supported by PNRR-PE-AI FAIR project funded by the NextGeneration EU program. We would like to thank David Lopez-Paz, Ricky Chen, Tianhong Li, and Yaron Lipman for fruitful discussions.
\else

\fi

\ifmeta
\else
\section*{LLM Usage Statement}
We used LLMs (Claude Opus, Gemini and GPT) to help rephrase and format the text of the paper.
In the work, we used generative AI (1)~to generate synthetic text captions for T2I training (Qwen3-VL-30B-A3B-Instruct and Llama-3.2-Vision), (2)~to reformat and package the dataset and make the data pipeline as efficient as possible, and to design or provide feedback on research methodology or experiments.
We have not used generative AI tools to help develop theoretical models or conceptual frameworks, or to propose or refine hypotheses.
Additionally, we used AI to assist us in coding, and to polish figures and plots and outline parts of the supplementary material.
We have reviewed all AI-assisted work and take responsibility for the final content of this work.
\fi

\ifmeta
\bibliographystyle{plainnat}
\bibliography{references}
\else
{\small
\bibliographystyle{ieeenat_fullname}
\bibliography{references}
}
\fi

\ifmeta\else
\clearpage
\onecolumn  
\setcounter{page}{1}
{\centering\Large\textbf{\thetitle}\\\vspace{0.5em}Supplementary Material\\\vspace{1.0em}\par}
\fi
\ifmeta\clearpage\fi  

\appendix

\startcontents[appendices] 
\printcontents[appendices]{}{1}{\section*{Contents of \ifmeta Appendix\else Supplementary Material\fi}} 

\section{Implementation details}
\label{app:pseudocode}

\subsection{Training and inference}

\paragraph{Training.}
Unless otherwise noted, we follow the JiT~\citep{Li2025} training recipe verbatim, apart from a slightly lower second EMA decay (0.999 instead of 0.9996) and changing only the items required by our wavelet formulation. Concretely: 600 epochs, 5 warmup epochs, AdamW ($\beta_1, \beta_2 = 0.9, 0.95$), batch size 1024, learning rate $2\!\times\!10^{-4}$ with constant schedule, weight decay 0, two EMAs with decays 0.9999 and 0.999, logit-normal time sampler $\mathrm{logit}(t) \sim \mathcal{N}(\mu, \sigma^2)$ with $\mu = -0.8, \sigma = 0.8$, noise scale $\sigma_\text{LF}$ for the LF band ($\text{res}/512$ for OpenImages and text-to-image; $1.5\cdot\text{res}/512$ for ImageNet, i.e.\ $0.75$ at $256{\times}256$ and $1.5$ at $512{\times}512$) and $\sigma_\text{HF} = 1$ for the HF bands, $(1-t)$-division clip of $0.05$, and class-token drop probability $0.1$ for CFG.

\paragraph{LF noise scale on ImageNet.}\label{app:noise_scale}
On OpenImages and in our text-to-image experiments we use $\sigma_\text{LF} = \text{res}/512$, i.e.\ JiT's noise scale evaluated at the LF resolution. On ImageNet we found a larger LF noise scale, $\sigma_\text{LF} = 1.5\cdot\text{res}/512$ ($0.75$ at $256{\times}256$ and $1.5$ at $512{\times}512$), to be crucial: with $\sigma_\text{LF} = \text{res}/512$ and fully re-tuned sampling (CFG scale and interval), FID is clearly worse for the B models than for the L models, and the B models are more sensitive to $\sigma_\text{LF}$ ($\sigma_\text{LF} = \text{res}/512$ vs.\ $1.5\cdot\text{res}/512$; B/16 at $256{\times}256$: 3.86 vs.\ 3.57; B/16 at $512{\times}512$: 3.15 vs.\ 2.96; B/32: 4.84 vs.\ 4.21; L/32: 2.70 vs.\ 2.76; L/16: 1.79 vs.\ 1.85). We attribute this to the power spectrum of ImageNet images: many are not natively high-resolution and have been upsampled or improperly downsampled, which distorts their frequency content relative to natural images. For datasets whose images are natively large and properly downsampled and cropped, such as our OpenImages subsets and the text-to-image data, we expect this multiplier to be unnecessary; there $\sigma_\text{LF} = \text{res}/512$ works well, as our OpenImages and text-to-image results show.

\paragraph{Video preprocessing.}
For video experiments, frames are decoded at a target sampling rate of 8\,FPS with a random temporal offset during training. Each frame is center-cropped to a square using the shorter spatial side and then resized to the target resolution ($128{\times}128$, or $256{\times}256$ for the Kinetics-600 XL/32 models) via bilinear interpolation with antialiasing. A random horizontal flip is applied during training.

\paragraph{Inference.}
For class- and text-conditioned image sampling we again follow the JiT setup: Heun ODE solver with 50 steps split between the two phases by $\rho$ and uniformly spaced within each (\Cref{app:step_allocation}), and CFG interval $[t_\text{cfg}, 1]$, where the interval start $t_\text{cfg}\in\{0.1, 0.15, 0.2, 0.25\}$ is tuned per model together with the CFG scale (JiT uses $[0.1, 1]$; with our two-phase schedule a later start, $0.2$--$0.25$, is consistently better). The CFG scale is swept in $[2.0, 4.5]$ with increment $0.25$--$0.5$ for class-conditional image generation, and in $[1.0, 10.0]$ with increment $0.5$ for text-to-image generation. The only parameter specific to our two-phase schedule, the Phase~0 step fraction $\rho$ (Section~\ref{sec:sampling}), is tuned on a grid of $0.05$ (values in \Cref{app:step_allocation}).

\paragraph{Model configurations.}
We use the standard JiT/DiT family of transformer backbones across all model scales. Architectural hyperparameters are summarized in \Cref{tab:model_configs}. WaiT inherits the exact same backbone for each scale and adds only the resolution embedding and stride-2 subsampled positional encodings for the coarse phase (\Cref{app:pseudocode}); the parameter counts therefore match JiT up to ${\leq}0.6\%$.

\begin{table}[htbp]
\centering
\normalsize
\setlength{\tabcolsep}{12pt}
\renewcommand{\arraystretch}{1.3}
\begin{tabular}{@{}l ccccc@{}}
\toprule
\textbf{Hyperparameter} & \textbf{B} & \textbf{L} & \textbf{XL} & \textbf{H} & \textbf{G} \\
\midrule
Depth                            & 12   & 24   & 28   & 32   & 40   \\
Hidden dim                       & 768  & 1024 & 1152 & 1280 & 1664 \\
Heads                            & 12   & 16   & 16   & 16   & 16   \\
Head dim                         & 64   & 64   & 72   & 80   & 104  \\
Bottleneck dim                   & 128  & 128  & 192  & 256  & 256  \\
Dropout (attn / proj)            & 0.0  & 0.0  & 0.0  & 0.2  & 0.2  \\
In-context start block           & 4    & 8    & 8    & 10   & 10   \\
\midrule
Parameters                       & ${\sim}$133M & ${\sim}$462M & ${\sim}$678M & ${\sim}$956M & ${\sim}$2B \\
\bottomrule
\end{tabular}
\vspace{4mm}
\caption{\textbf{Model configurations} for JiT and WaiT backbones at scales B/L/XL/H/G. All variants use in-context length $=32$ class tokens (when in-context conditioning is enabled). Image size is $256{\times}256$ (or $512{\times}512$/$1024{\times}1024$ depending on the experiment).}
\label{tab:model_configs}
\end{table}

\paragraph{Pseudo code.}
For reproducibility, we provide pseudocode for both the standard JiT baseline and our Wavelet-aware image Transformer extension, side by side, for both training in \Cref{tab:pseudocode_train} and sampling in \Cref{tab:pseudocode_sample}. In both tables, the last argument of $f_\theta$, $\texttt{res} \in \{0, 1\}$, indicates the coarse or fine phase. The model maps it to the side length of the corresponding token grid (e.g., 8 and 16 for a /32 model at $512{\times}512$), which is encoded with a sinusoidal embedding and MLP and added to the time and class conditioning. In the coarse phase, the fixed sin-cos positional embeddings and the RoPE positions are those of the full-resolution grid subsampled with stride 2, so each coarse token carries the position of the image region it covers.

\begin{table}[htbp]
\centering
\small
\setlength{\tabcolsep}{6pt}
\begin{tabular}{p{0.46\textwidth}|p{0.46\textwidth}}
\toprule
\textbf{Standard JiT (Baseline)} & \textbf{Wavelet-aware image Transformer (Ours)} \\
\midrule
\textbf{Input:} image $x$, label $y$ &
\textbf{Input:} image $x$, label $y$, threshold $t^*$, scale $S_\text{LF}$ \\[4pt]

1. Sample $t \sim \mathcal{LN}(0, 1)$ &
1. Sample $t_\text{LF} \sim \mathcal{LN}(0, 1)$ \\[2pt]

2. $\varepsilon \sim \mathcal{N}(0, \sigma^2 I)$ &
2. {$x_\text{LF}, x_\text{HF} = \text{DWT}(x)$} \\[2pt]

3. $z = t \cdot x + (1{-}t) \cdot \varepsilon$ &
3. {$\tilde{x}_\text{LF} = x_\text{LF} \,/\, S_\text{LF}$} \\[4pt]

& {\textit{--- Phase 0 (coarse, $H/2 \times W/2$) ---}} \\[2pt]

4. $\hat{x} = f_\theta(z, t, y)$ &
4. {$\varepsilon_c \sim \mathcal{N}(0, \sigma_\text{LF}^2 I)$} \\[2pt]

5. $v = (x - z) / (1{-}t)$ &
5. {$z_c = t_\text{LF} \cdot \tilde{x}_\text{LF} + (1{-}t_\text{LF}) \cdot \varepsilon_c$} \\[2pt]

6. $\hat{v} = (\hat{x} - z) / (1{-}t)$ &
6. {$\hat{\tilde{x}}_\text{LF}$$ = f_\theta(z_c, t_\text{LF}, y, \texttt{res}=0)$} \\[2pt]

7. $\mathcal{L} = \|v - \hat{v}\|^2$ &
7. {$\mathcal{L}_c = \|$$\hat{\tilde{x}}_\text{LF}$$ - \tilde{x}_\text{LF}\|^2 / (1{-}t_\text{LF})^2$} \\[4pt]

& {\textit{--- Phase 1 (fine, $H \times W$) ---}} \\[2pt]
& 8. {Sample $t'_\text{LF} \sim \mathcal{LN}(t^*, 1)$} \\[2pt]
& 9. {$\varepsilon_\text{LF} \sim \mathcal{N}(0, \sigma_\text{LF}^2 I)$, \; $\varepsilon_\text{HF} \sim \mathcal{N}(0, I)$} \\[2pt]
& 10. {$z_\text{LF} = t'_\text{LF} \cdot \tilde{x}_\text{LF} + (1{-}t'_\text{LF}) \cdot \varepsilon_\text{LF}$} \\[2pt]
& \quad {$t_\text{HF} = (t'_\text{LF}{-}t^*)/(1{-}t^*)$} \\[2pt]
& 11. {$z_\text{HF} = t_\text{HF} \cdot x_\text{HF} + (1{-}t_\text{HF}) \cdot \varepsilon_\text{HF}$} \\[2pt]
& 12. {$z_f = \text{IDWT}(z_\text{LF} \cdot S_\text{LF},\; z_\text{HF})$} \\[2pt]
& 13. {$\hat{x} = f_\theta(z_f, t'_\text{LF}, y, \texttt{res}=1)$} \\[2pt]
& 14. {$(\hat{x}_\text{LF}, \hat{x}_\text{HF}) = \text{DWT}(\hat{x})$} \\[2pt]
& 15. {$\mathcal{L}_f = \dfrac{\|\hat{x}_\text{LF}/S_\text{LF} - \tilde{x}_\text{LF}\|^2}{(1{-}t'_\text{LF})^2} + \dfrac{\|\hat{x}_\text{HF} - x_\text{HF}\|^2}{(1{-}t_\text{HF})^2}$} \\[4pt]

\textbf{Return:} $\mathcal{L}$ &
\textbf{Return:} $\mathcal{L}_c$ (Phase 0 batch) or $\mathcal{L}_f$ (Phase 1 batch) \\
\bottomrule
\medskip
\end{tabular}
\caption{\textbf{Training pseudocode comparison.} Left: standard flow-matching (JiT baseline). Right: our Wavelet-aware image Transformer extension. $\mathcal{LN}(a, b)$: logit-normal distribution, $\mathrm{logit}(u) \sim \mathcal{N}(\mu, \sigma^2)$ with $\mu = -0.8$, $\sigma = 0.8$, rescaled to $[a, b]$ as $t = a + (b - a)\,u$. $S_\text{LF}$: 95th-percentile normalization constant (Section~\ref{sec:normalization}). $\varepsilon_c, \varepsilon_\text{LF}, \varepsilon_\text{HF}$: independent Gaussian noise samples. $\sigma_\text{LF} = \text{image size}/512$: LF noise scale (the JiT noise scale at the LF resolution, multiplied by 1.5 on ImageNet as explained in \Cref{app:noise_scale}). Each squared norm is averaged over the coefficients of its band, so the LF and HF terms of $\mathcal{L}_f$ contribute equally despite their different sizes ($D/4$ vs.\ $3D/4$). \ifmeta The full implementation is available at \url{https://github.com/facebookresearch/WaiT}.\else Code will be made publicly available upon acceptance.\fi}
\label{tab:pseudocode_train}
\end{table}

\begin{table}[htbp]
\centering
\small
\setlength{\tabcolsep}{6pt}
\begin{tabular}{p{0.46\textwidth}|p{0.46\textwidth}}
\toprule
\textbf{Standard JiT (Baseline)} & \textbf{Wavelet-aware image Transformer (Ours)} \\
\midrule
\textbf{Input:} labels $y$, steps $K$, CFG $w$ &
\textbf{Input:} labels $y$, steps $K$, CFG $w$, $t^*$, Phase~0 fraction $\rho$, scale $S_\text{LF}$ \\[4pt]

& {\textit{--- Step schedule ---}} \\[2pt]
1. $\{t_i\} = \text{linspace}(0, 1, K{+}1)$ &
1. $K_1 = \mathrm{round}(\rho K)$, \; $K_2 = K - K_1$ \\[2pt]
& 2. $\{t_i\} = \text{linspace}(0, t^*, K_1{+}1) \cup \text{linspace}(t^*, 1, K_2{+}1)$ \\[4pt]

2. $z \sim \mathcal{N}(0, \sigma^2 I)$ \quad [$H \times W$] &
3. {$z \sim \mathcal{N}(0, \sigma_\text{LF}^2 I)$} \quad [{$H/2 \times W/2$}] \\[4pt]

& {\textit{--- Phase 0 ($t{=}0 \to t^*$, coarse, normalized LF only) ---}} \\[2pt]
3. \textbf{for} $i = 0$ to $K{-}1$: &
4. {\textbf{for} $i = 0$ to $K_1{-}1$:} \\[2pt]
\quad $\hat{x} = f_\theta(z,t_i,\emptyset) + w\,[f_\theta(z,t_i,y) - f_\theta(z,t_i,\emptyset)]$ &
\quad $\hat{\tilde{x}}_\text{LF}$$ = f_\theta(z, t_i, \emptyset, 0) + w\,[f_\theta(z, t_i, y, 0) - f_\theta(z, t_i, \emptyset, 0)]$ \\[2pt]
\quad $v = (\hat{x} - z)/(1{-}t_i)$ &
\quad {$v = ($$\hat{\tilde{x}}_\text{LF}$$ - z)/(1{-}t_i)$} \\[2pt]
\quad $z \leftarrow z + (t_{i+1} - t_i) \cdot v$ &
\quad $z \leftarrow z + (t_{i+1} - t_i) \cdot v$ \\[4pt]

& {\textit{--- Transition at $t^*$: un-norm. LF, inject HF noise ---}} \\[2pt]
& 5. {$z_\text{LF} = z \cdot S_\text{LF}$} \\[2pt]
& 6. {$z_\text{HF} \sim \mathcal{N}(0, I)$ \quad [HF resolution]} \\[2pt]
& 7. {$z = \text{IDWT}(z_\text{LF}, z_\text{HF})$} \quad [$H \times W$, pixel space] \\[4pt]

& {\textit{--- Phase 1 ($t^* \to 1$, fine, LF\,+\,HF in pixel space) ---}} \\[2pt]
& 8. {\textbf{for} $i = K_1$ to $K{-}1$:} \\[2pt]
& \quad $\hat{x} = f_\theta(z, t_i, \emptyset, 1) + w\,[f_\theta(z, t_i, y, 1) - f_\theta(z, t_i, \emptyset, 1)]$ \\[2pt]
& \quad $v = (\hat{x} - z)/(1{-}t_i)$ \\[2pt]
& \quad $z \leftarrow z + (t_{i+1} - t_i) \cdot v$ \\[4pt]

\textbf{Return:} $z$ &
\textbf{Return:} $z$ \\
\bottomrule
\end{tabular}
\caption{\textbf{Sampling pseudocode comparison.} Left: standard single-phase sampling. Right: our two-phase cascade. The model performs $x$-prediction internally. Phase~0 operates on $4\times$ fewer tokens at $H/2 \times W/2$ resolution. $\sigma_\text{LF}$ is the LF noise scale defined in \Cref{tab:pseudocode_train}. The transition at $t^*$ un-normalizes the LF state by $S_\text{LF}$ and re-initializes the HF bands as fresh $\mathcal{N}(0, I)$ noise (matching the distribution of $\varepsilon_\text{HF}$ at training time, which the model expects to see when $t_\text{HF}=0$). Heun steps can replace Euler (doubles NFE). \ifmeta The full implementation is available at \url{https://github.com/facebookresearch/WaiT}.\else Code will be made publicly available upon acceptance.\fi}
\label{tab:pseudocode_sample}
\end{table}

\subsection{Step Allocation}
\label{app:step_allocation}

Given a budget of $K$ ODE steps and a Phase~0 step fraction $\rho \in (0, 1)$, the sampler uses $K_1 = \mathrm{round}(\rho K)$ uniformly spaced steps on $[0, t^*]$ for the coarse Phase~0 and $K_2 = K - K_1$ uniformly spaced steps on $[t^*, 1]$ for the fine Phase~1. Because each Phase~0 step processes $4\times$ fewer tokens, $\rho$ directly controls the compute--quality trade-off.

\paragraph{Values used in our experiments.} Unless otherwise noted we use $t^* = 0.25$ and $K = 50$ throughout. For the ImageNet $512{\times}512$ models in \Cref{tab:pixel_comparison} we use $\rho = 0.6$ for B/32, B/16, G/32 and G/16, $\rho = 0.5$ for L/32, $\rho = 0.55$ for H/32, and $\rho = 0.65$ for L/16 and H/16 (per-model settings in \Cref{app:per_model}). In all models FID varies little for $\rho \le 0.6$ and degrades sharply beyond $0.7$.

\paragraph{Per-model settings.}\label{app:per_model}
\Cref{tab:per_model} lists the sampling settings of the ImageNet $512{\times}512$ models in \Cref{tab:pixel_comparison}; all use $\sigma_\text{LF} = 1.5$, $t^*=0.25$, $K=50$ Heun steps and 600 training epochs. Each row is a single sampling configuration and a single EMA for all three metrics.

\begin{table}[h]
\centering
\small
\caption{Per-model settings (ImageNet $512{\times}512$).}
\label{tab:per_model}
\begin{tabular}{@{}lccccc@{}}
\toprule
Model & $\rho$ & CFG interval & CFG & EMA & GFLOPs \\
\midrule
WaiT-B/32 & 0.6 & $[0.25, 1]$ & 4.5 & 0.999 & 15 \\
WaiT-B/16 & 0.6 & $[0.25, 1]$ & 3.5 & 0.9999 & 64 \\
WaiT-L/32 & 0.5 & $[0.2, 1]$ & 3.0 & 0.999 & 58 \\
WaiT-L/16 & 0.65 & $[0.2, 1]$ & 3.0 & 0.999 & 201 \\
WaiT-H/32 & 0.55 & $[0.2, 1]$ & 3.0 & 0.999 & 113 \\
WaiT-H/16 & 0.65 & $[0.25, 1]$ & 3.5 & 0.999 & 411 \\
WaiT-G/32 & 0.6 & $[0.2, 1]$ & 3.0 & 0.999 & 225 \\
WaiT-G/16 & 0.6 & $[0.2, 1]$ & 3.0 & 0.999 & 914 \\
\bottomrule
\end{tabular}
\end{table}

\subsection{Band Normalization}
\label{app:normalization}

This section provides additional details on the normalization scheme summarized in Section~\ref{sec:normalization}.
A practical but critical detail for stable training is the normalization of wavelet bands. The DWT redistributes signal energy unevenly across bands: for the orthonormal Haar wavelet, the LF (low-frequency) band computes the energy-normalized sum of $2\times 2$ pixel blocks, producing values in $[\text{-}2, 2]$ when the input is normalized to $[\text{-}1, 1]$. The HF detail bands, by contrast, capture differences and are naturally small-valued (typical standard deviation ${\sim}0.05$--$0.19$). This dynamic range mismatch means the flow-matching interpolation $z_t = t \cdot x + (1-t) \cdot \epsilon$ operates in different regimes for each band if left unnormalized.

\paragraph{Choice of normalization constant.} We use the \textbf{95th percentile} ($p_{95}$) of the absolute LF coefficient values, computed empirically over the training set:
$$ S_{LF} = \text{Quantile}_{0.95}\big(\,|x_{LF}|\,\big), $$
where the quantile is taken over all spatial positions and all training images. This is computed once per dataset as a preprocessing step. We prefer $p_{95}$ over the maximum because the maximum is sensitive to rare outlier images, leading to an overly conservative scale that compresses the effective dynamic range. The $p_{95}$ percentile provides a robust estimate that maps the vast majority of LF values to $[\text{-}1, 1]$, with only 5\% of values slightly exceeding this range, a negligible effect that the network learns to handle. Compared to standard deviation normalization, $p_{95}$ better preserves the bounded structure of the data distribution, avoiding the heavy tails that arise from std-based scaling.

\paragraph{Dataset-specific values.} In practice, we pre-compute $S_{LF}$ for each (dataset, resolution, wavelet filter) combination. For example, $S_{LF} = 1.94$ for ImageNet 256$\times$256 with the Haar wavelet, and $S_{LF} = 1.95$ for OpenImages 512$\times$512 (the normalization constant is dataset-specific as the frequency content distribution varies across datasets). However, $S_{LF}$ is remarkably stable across datasets of similar content, and the method is not sensitive to small perturbations in its value.

This formulation yields a remarkably simple recipe: replace standard flow-matching with band-specific schedules in wavelet space, normalize the LF band by its empirical $p_{95}$, and inherit the entire unchanged JiT training and sampling pipeline.

In our preliminary analysis we experimented with applying an analogous per-band normalization to the HF coefficients, but did not observe any consistent benefit on the datasets considered here. We attribute this to the relatively small dynamic range and well-behaved sparsity of HF coefficients in natural images at the resolutions we study, which already match the unit-variance noise schedule reasonably well. We expect this picture may change for substantially larger or more diverse datasets whose high-frequency power spectra and tail behavior differ markedly from natural images (e.g., synthetic, scientific, or highly textured imagery), where an HF normalization analogous to $S_{LF}$ could yield further stability and quality gains.

\paragraph{Ablation of the normalization scheme.}
We compare LF normalization choices---none, $p_{95}$/$p_{99}$ percentile, maximum, and standard-deviation scaling (all computed globally over the training set)---across pixel and latent spaces, multiple datasets, and up to three DWT levels (\Cref{tab:norm_ablation}). Percentile scaling ($p_{95}$/$p_{99}$) is consistently best; removing normalization degrades performance in every setting, and standard-deviation scaling is worst (its heavy tails skew the variance). We therefore adopt $p_{95}$. The strong sensitivity in the latent-space (DDT) and $3$-level settings confirms that normalization is a fundamental requirement across architectures, not a pixel-space-specific detail.

\begin{table}[h]
\centering
\small
\caption{\textbf{Ablation of the LF normalization scheme} (FID-50k, lower is better; $p_{95}$, ours, in bold).}
\label{tab:norm_ablation}
\begin{tabular}{@{}lccccc@{}}
\toprule
Setting & No norm. & $p_{95}$ & $p_{99}$ & Max & Std. \\
\midrule
ImageNet 512, pixel, B/16          & 3.89  & \textbf{2.96}  & 3.01  & 3.05  & 4.72 \\
ImageNet 512, latent DDT           & 3.53  & \textbf{1.33}  & 1.37  & 1.36  & 5.71 \\
OpenImages 512, two levels, L/16    & 17.98 & \textbf{15.10} & \textbf{15.10} & 15.23 & 20.32 \\
OpenImages 1024, three levels, L/32 & 16.14 & \textbf{10.11} & 10.32 & 10.72 & 27.12 \\
\bottomrule
\end{tabular}
\end{table}

\ifmeta  
\subsection{Code and model attribution}
\label{app:attribution}

\paragraph{JiT.} This work builds on the Just image Transformers (JiT) framework of \citet{Li2025}. Our implementation extends their publicly released codebase (MIT license), from which we adopt the model architecture, the training recipe and the $x$-prediction formulation with $v$-loss; our wavelet-aware modifications are applied on top. For the JiT/32 rows in \Cref{tab:pixel_comparison} we evaluate their released weights, and the JiT/16 models we train ourselves with their code.

\paragraph{Baselines.} For the 5-crop FID and hFWD of the baselines in \Cref{tab:pixel_comparison}, and for the qualitative comparisons, we run the released pretrained models with the following codebases:
\begin{itemize}[leftmargin=1.4em,topsep=2pt,itemsep=1pt]
  \item DiT-XL/2~\citep{peebles23iccv}\ifmeta: \url{https://github.com/facebookresearch/DiT}\fi\ (CC-BY-NC 4.0)
  \item DDT-XL/2~\citep{wang2025ddt}\ifmeta: \url{https://github.com/MCG-NJU/DDT}\fi
  \item RAE~\citep{zheng202rae}\ifmeta: \url{https://github.com/bytetriper/RAE}\fi\ (MIT)
  \item ADM-G~\citep{dhariwal21nips}\ifmeta: \url{https://github.com/openai/guided-diffusion}\fi\ (MIT)
  \item PixelDiT~\citep{yu2025pixeldit}\ifmeta: \url{https://github.com/NVlabs/PixelDiT}\fi\ (NVIDIA Source Code License)
\end{itemize}
The latent-space experiment (\Cref{tab:latent}) is additionally implemented on top of the DDT codebase. We thank the authors of all these works for making their code and models publicly available.
\fi

\section{Mutual Information Estimation}
\label{app:mi_estimation}

The mutual information curves in \Cref{fig:mi_comparison} measure, for a single time step $t$, how much information the noisy wavelet coefficients carry about the clean ones. We use a simple, model-agnostic plug-in histogram estimator.

\paragraph{Wavelet decomposition.} For an image (or a state along a generative trajectory) we apply a single-level orthonormal Haar DWT, which yields one coarse band $c \in \mathbb{R}^{D/4}$ (LF) and three detail bands stacked into a single fine vector $f = [\text{LH};\text{HL};\text{HH}] \in \mathbb{R}^{3D/4}$. We work directly with the flattened coefficient vectors, treating the different HF channels as independent draws of the same random variable.

\paragraph{Plug-in MI estimator.} Given two scalar samples $X = \{x_i\}_{i=1}^N$ and $Y = \{y_i\}_{i=1}^N$ obtained by concatenating coefficients across many spatial positions (and, for the backward setting, across many trajectories), we form a 2D histogram with $K = 64$ uniform bins per axis to obtain a discrete joint distribution $\hat{p}(x, y)$ together with the marginals $\hat{p}(x), \hat{p}(y)$.
The mutual information is then estimated as
\[
\widehat{\mathrm{MI}}(X, Y) \;=\; \sum_{x, y \,:\, \hat{p}(x,y) > 0} \hat{p}(x, y) \log \frac{\hat{p}(x, y)}{\hat{p}(x)\,\hat{p}(y)}.
\]
We apply this estimator independently to the coarse pair $(c_t,c_1)$ and the fine pair $(f_t, f_1)$, where $c_1, f_1$ are the clean (or final) wavelet coefficients and $c_t, f_t$ are the noisy ones at step $t$.

\paragraph{Forward (real images, top panel).} We sample $\epsilon \sim \mathcal{N}(0, I)$, form $x_t = t\,x_1 + (1-t)\epsilon$, take the Haar DWT, and compute $\widehat{\mathrm{MI}}(c_t, c_1)$ and $\widehat{\mathrm{MI}}(f_t, f_1)$ at a uniform grid of 41 timesteps in $[0,1]$. Here $x_1$ are $50$ real images randomly sampled from ImageNet at $256{\times}256$ resolution, and we average each MI over these images and over independent noise draws to reduce variance.

\paragraph{Backward (generated images, bottom panel ).} For JiT, we cache the intermediate trajectory states across a large set of generated images. For each saved timestep we treat the model's final sample (last step) as $x_1$ and the current step as $x_t$, decompose both, and concatenate coefficients across all images before computing $\widehat{\mathrm{MI}}(c_1, c_t)$ and $\widehat{\mathrm{MI}}(f_1, f_t)$. This pools the empirical joint distribution over many spatial positions and many trajectories, yielding a single MI value per band per step. Concretely, the curves in the bottom panel of \Cref{fig:mi_comparison} are obtained by averaging over 50 generated samples of the \emph{golden retriever} class from a JiT-B/32 model trained on ImageNet $512{\times}512$.

\paragraph{Crossover threshold.} In both panels the dashed vertical line and star mark where the fine-band MI drops below $0.01$ nats.

\section{Evaluation metrics: 5-crop FID and high-frequency FWD}
\label{app:metrics}

Both metrics are designed to evaluate generation quality at native resolution, avoiding the aggressive $299{\times}299$ downsampling of standard FID that destroys exactly the high-frequency content WaiT is built to model.

\paragraph{5-crop FID (5cFID).} Instead of resizing each image, we extract five $299{\times}299$ patches at the native resolution -- the four corners and the center -- and pass each patch through InceptionV3 to obtain its $2048$-d feature. Features from all crops of all generated images are pooled into a single set, from which we estimate the Gaussian mean and full covariance and compute the standard Fr\'echet distance against precomputed reference statistics for the matching (dataset, resolution, crop strategy), computed on the full training set. We applied the same protocol to alternative crop variants (center-only, four-corner, and a random crop with a fixed seed) and found the relative ranking of methods to be unchanged across all of them; we therefore report only the 5-crop variant in the paper.

\paragraph{High-frequency FWD (hFWD).} Building on the Fr\'echet Wavelet Distance~\citep{veeramacheneni2025fwd}, hFWD replaces InceptionV3 features with a level-$4$ Haar wavelet packet transform, which splits each image into $4^4 = 256$ packets: packet $0$ is the pure low-pass (DC) packet, and the remaining $255$ cover all other combinations of low- and high-pass filtering. Each packet's coefficients are flattened into one descriptor ($3\times32\times32 = 3072$ values at $512{\times}512$; larger packets, e.g.\ at $1024{\times}1024$, are first average-pooled to $32\times32$), whose mean and covariance we estimate from first and second moments. hFWD is the Fr\'echet distance between the generated and reference statistics (the latter computed on the full training set at the same resolution), averaged over the $255$ non-DC packets; we drop the DC packet to isolate texture fidelity (as low-frequencies are already captured by FID).

\section{Additional  results}
\label{app:qualitative}
\label{app:imagenet_qualitative}

\subsection{Additional Ablations in ImageNet 256 resolution}
\label{app:ablations}

Here we provide  ablation results summarized in Fig.~\ref{tab:ablation} of the main text.
For these ablations  we  use WaiT-B/16 models trained on ImageNet 256 resolution for 1k epochs, to ensure all models have converged in terms of FID.

In \Cref{tab:schedules_app} we compare the different noise schedules visualized in \Cref{fig:schedules}.
Our \textit{Delayed Linear} schedule eliminates the train-test discrepancy, achieving the best balance between FID and efficiency. When tuned, the Discontinuous schedule reaches a slightly lower FID ($3.41$ vs.\ $3.57$), but at $10\%$ more compute ($70\%$ vs.\ $60\%$ of JiT GFLOPs).
Note that, unlike the \emph{cumulative} ablation in \Cref{tab:ablation} of the main text---which starts from a naïve two-stage baseline ($t^*{=}0.5$, coarse trained only on $[0,t^*]$) and adds one improvement per row---each schedule in \Cref{tab:schedules_app} is reported with its own \emph{tuned} setting (best $t^*$ and global coarse training). For our Delayed Linear row (WaiT-B/16 at $256{\times}256$) we use $t^*=0.25$, $\sigma_\text{LF} = 0.75$ ($=1.5\cdot\text{res}/512$), $\sigma_\text{HF}=1$ and $S_\text{LF}=1.94$, train for 1000 epochs with batch size 1024, learning rate $2\times10^{-4}$ (constant after 5 warmup epochs) and EMA decay 0.9999, and sample with 50 Heun steps, $\rho=0.6$ (${\approx}60\%$ of JiT's GFLOPs), CFG interval $[0.25, 1]$ and CFG scale 4.0.

\begin{figure}[H]
    \centering
    \includegraphics[width=\linewidth]{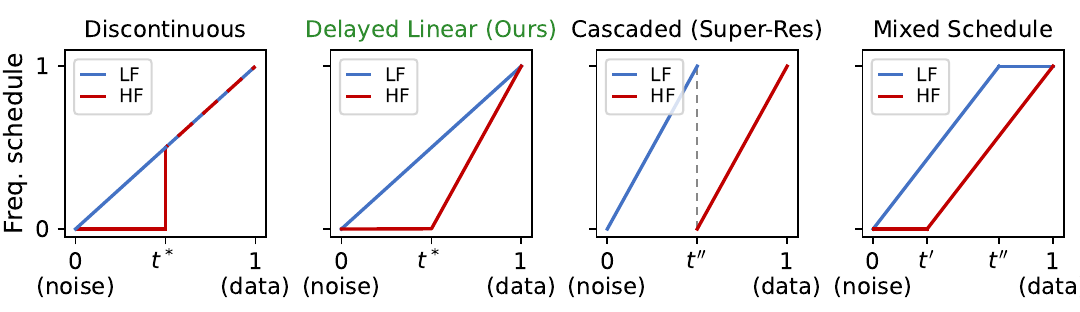}
    \caption{\textbf{Multi-resolution schedule variants compared in \Cref{tab:schedules_app}.} Blue: LF band signal coefficient; Red: HF band signal coefficient. All three tested schedules have HF band being exactly pure noise at $t^*$, removing the train--test discontinuity present in the classic Discontinuous schedule.}
    \label{fig:schedules_app}
\end{figure}

On top of the two schedules already discussed in the main text (\textit{Discontinuous} and \textit{Delayed Linear}), we also experimented with two additional variants: \textit{Cascaded (super-resolution)} and \textit{Mixed} which we report here for completeness. We briefly describe all four schedule variants below; they are visualized in \Cref{fig:schedules_app}:
\begin{itemize}[leftmargin=1.2em,topsep=2pt,itemsep=1pt]
    \item \textbf{Discontinuous}: A naive two-stage cascade in the spirit of PixelFlow~\citep{chen2025pixelflow} and Pyramidal Flow~\citep{jin2024pyramidalflowmatchingefficient}. The LF band is denoised on $[0, t^*]$, then at $t^*$ HF noise is injected and both bands are jointly denoised on $[t^*, 1]$. The hard handoff at $t^*$ creates a train--test mismatch (the HF band is unseen at exactly the noise level injected at inference).
    \item \textbf{Delayed Linear (ours)}: The LF band follows the standard linear schedule on $[0, 1]$, while the HF band stays as pure noise until $t^*$ and then linearly interpolates from noise to data on $[t^*, 1]$. By construction the HF distribution at $t^*$ is exactly $\mathcal{N}(0, I)$, eliminating the discontinuity.
    \item \textbf{Cascaded (super-resolution)}: A strictly sequential cascade with a single model: the LF band is first denoised all the way to a clean image, i.e.\ until $t_{LF}=1$, which is reached at time $t''$ of the shared schedule. Conditioned on this clean LF image, the same model then upsamples it and denoises the HF bands on $[t'', 1]$. HF therefore only starts once LF is fully clean, mirroring cascaded super-resolution diffusion~\citep{ho22cascaded}, but with one jointly trained network instead of separately trained models.
    \item \textbf{Mixed}: LF and HF follow parallel linear schedules but with different start times; LF begins at $t=0$ and reaches data at $t''$, while HF waits until $t'<t''$ and then climbs in parallel to LF, reaching data at $t=1$. The key difference from \emph{Cascaded} is the overlap window $[t', t'']$ during which \emph{both} bands are simultaneously partially noisy and are denoised \emph{jointly}, allowing HF generation to be conditioned on a still-noisy LF context (rather than waiting for a fully clean LF as in the Cascaded variant).
\end{itemize}
A figure visualizing all four schedules is provided in \Cref{fig:schedules_app}.

\begin{table}[H]
  \caption{\textbf{Comparison of multi-resolution schedules.}
  Evaluated on ImageNet 256$\times$256 using WaiT-B/16.
  GFLOPs and savings relative to JiT baseline.}
  \label{tab:schedules_app}
  \centering
  \small
  \begin{tabular}{@{}lccc@{}}
    \toprule
    Schedule & FID ($\downarrow$) & GFLOPs & Savings \\
    \midrule
    Baseline (JiT-B/16) & 3.81 & 100\% & 0\% \\
    \midrule
    Discontinuous & \bf 3.41 & 70\% & 30\% \\
    Cascaded & 4.50 & 75\% & 25\% \\
    Mixed & 4.20 & 75\% & 25\% \\
      \rowcolor{yellow!15}
  \textbf{Delayed linear (ours)} & 3.57 & \textbf{60\%} & \textbf{40\%} \\
    \bottomrule
  \end{tabular}
\end{table}

In our final ablation in \Cref{tab:interval_app} we consider whether the coarse training is performed over the full $[0,1]$ interval  or only from 0 to $t^*$.
We find that training the coarse stage over the full $[0,1]$ interval leads to best results, improving FID from 4.81 to 3.57 (at $t^*{=}0.25$; the corresponding $-1.10$ gain in \Cref{tab:ablation} of the main text is measured at $t^*{=}0.5$).
We speculate that exposing the model to LF tokens across the entire trajectory by including the regime $t > t^*$ where it is also conditioned on noisy HF inputs may help it implicitly learn the relationship between coarse and fine frequency content, rather than treating the LF band as an isolated subproblem.
\begin{table}[H]
  \caption{\textbf{Ablation of coarse training range.}
  Evaluated on ImageNet 256$\times$256 using WaiT-B/16 at $t^*=0.25$.  }
  \label{tab:interval_app}
  \centering
  \small
  \begin{tabular}{@{}lc@{}}
    \toprule
    Training Range (Coarse) & FID ($\downarrow$) \\
    \midrule
    Restricted $[0, t^*]$ & 4.81 \\
      \rowcolor{yellow!15}
  \textbf{Full $[0, 1]$ (ours)} & \textbf{3.57} \\
    \bottomrule
  \end{tabular}
\end{table}

\paragraph{Sensitivity to the wavelet family.}
\label{app:wavelet}
Here we ablate the wavelet basis on ImageNet-256 with WaiT-B/16 at $t^*{=}0.25$ (\Cref{tab:wavelet_ablation}). Performance is stable across families (within $0.19$ FID); we adopt Haar throughout for its simplicity---it is orthogonal, lossless, has minimal spatial support, and its LL band is a clean $2\times2$ average pool---and note that the strong results of wider-support bases (CDF~9/7, Symlet) show this image-like property is not required for WaiT to work.

\begin{table}[h]
\centering
\small
\caption{\textbf{Sensitivity to the wavelet family} (ImageNet-256, WaiT-B/16, $t^*{=}0.25$).}
\label{tab:wavelet_ablation}
\begin{tabular}{@{}lccccc@{}}
\toprule
Wavelet & Haar & CDF~5/3 & CDF~9/7 & Db2 & Symlet \\
\midrule
FID ($\downarrow$) & 3.57 & \textbf{3.48} & 3.53 & 3.62 & 3.67 \\
\bottomrule
\end{tabular}
\end{table}

\subsection{Perceptual validation of the evaluation metrics}
\label{app:perceptual}
To confirm that 5cFID and hFWD capture human perception, we evaluate them on the PIPAL benchmark~\citep{gu2020pipal}, which pairs ground-truth images with high-frequency distortions mirroring generative failure modes (super-resolution errors, blur, noise, GAN artifacts) and provides ${\sim}1.13$M human Elo judgments across $116$ distortion groups. Since 5cFID and hFWD are distribution-level distances, we evaluate them per distortion group: for each group we pool the features of all its distorted images and compute their Fr\'echet distance to the statistics of all clean PIPAL reference images, giving one 5cFID and one hFWD value per group. The human score of a group is the mean Elo of its distorted images. We then report the Pearson (PLCC, linear) and Spearman (SRCC, rank-order) correlations between the (negated) metric values and the mean Elo scores across all 116 groups (\Cref{tab:pipal}). Both show strong, statistically significant agreement (Holm--Bonferroni-corrected and permutation-tested, $p<4\times10^{-4}$), confirming that WaiT's metric gains reflect genuine perceptual improvements rather than overfitting to the wavelet domain.

\begin{table}[h]
\centering
\small
\caption{\textbf{Correlation of 5cFID and hFWD with human preference} on PIPAL.}
\label{tab:pipal}
\begin{tabular}{@{}lcccc@{}}
\toprule
Metric & $|\text{PLCC}|$ & PLCC $p$-value & $|\text{SRCC}|$ & SRCC $p$-value \\
\midrule
5cFID & 0.695 & $<4.7\times10^{-18}$ & 0.656 & $<1.38\times10^{-15}$ \\
hFWD  & \textbf{0.708} & $<6.69\times10^{-19}$ & \textbf{0.663} & $<5.15\times10^{-16}$ \\
\bottomrule
\end{tabular}
\end{table}

\subsection{ImageNet 512 resolution}

\begin{figure}
  \centering
  \ifmeta\includegraphics[width=\linewidth]{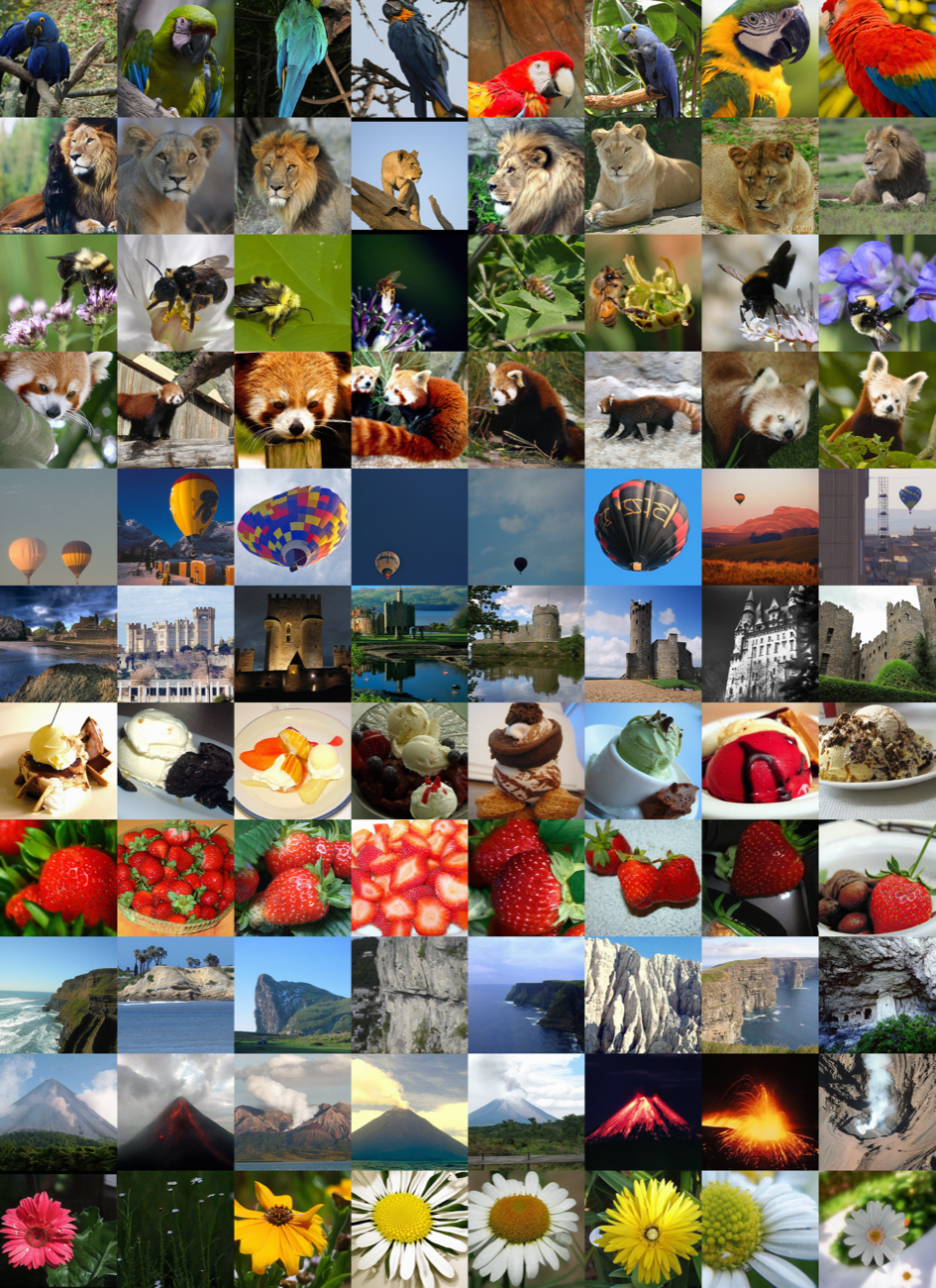}\else\includegraphics[width=\linewidth,height=0.86\textheight,keepaspectratio]{plots/bandlimited_generations.png}\fi
  \caption{\textbf{Uncurated class-conditional samples from WaiT-H/16, ImageNet 512$\times$512.}
  Each row shows eight samples from a single class. From top to bottom: \emph{macaw, lion, bee, red panda, balloon, castle, ice cream, strawberry, cliff, volcano, daisy}.}
  \label{fig:bandlimited_generations}
\end{figure}

In \Cref{fig:bandlimited_generations} we provide samples for ImageNet 512 resolution class-conditional generations.

\begin{figure}
  \centering
  \includegraphics[width=\linewidth]{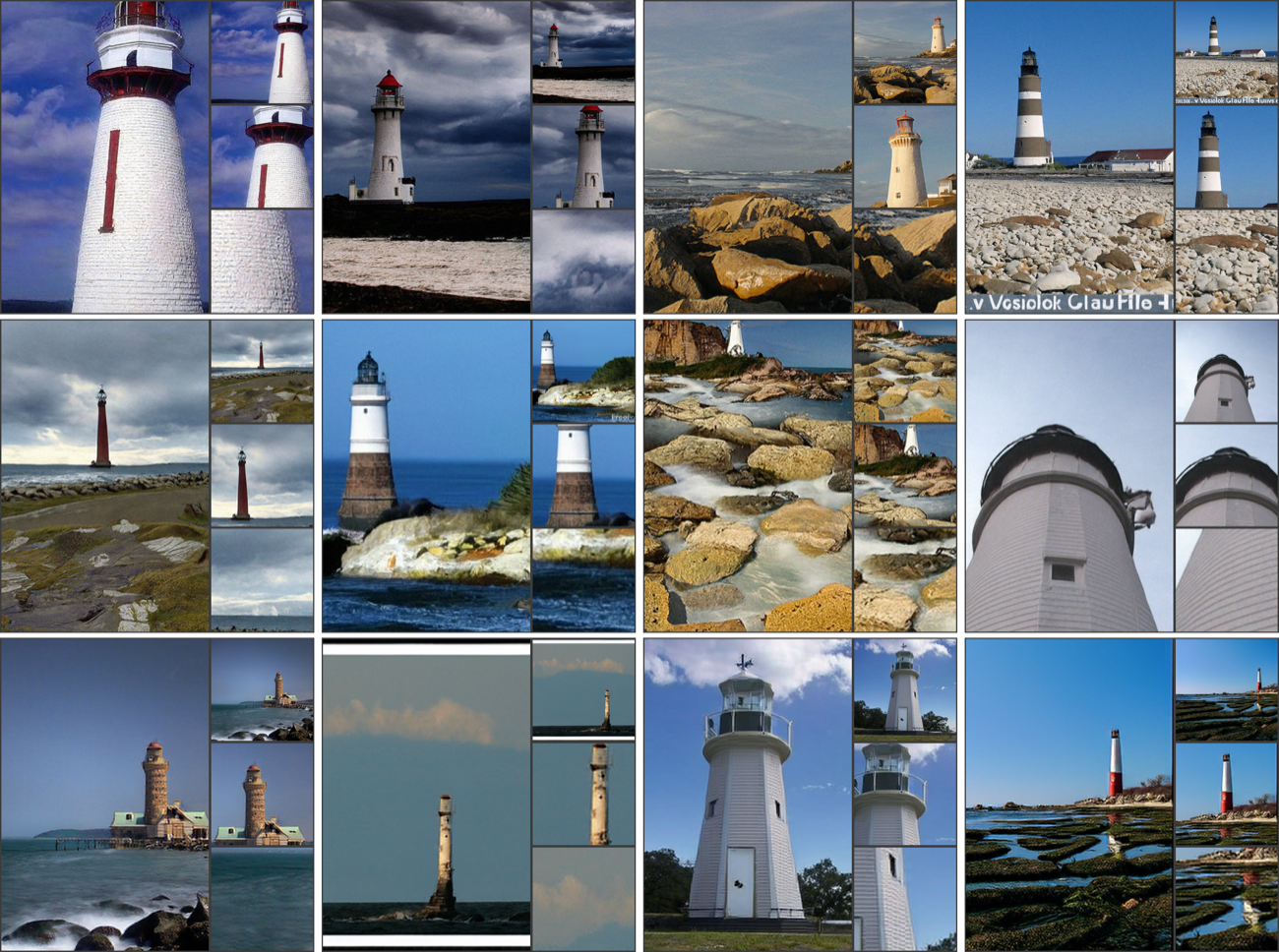}
  \caption{\textbf{Single-class comparison (lighthouse)} across methods on ImageNet 512$\times$512. \textbf{Top:} WaiT-H/32  (ours). \textbf{Middle:} JiT baseline. \textbf{Bottom:} RAE latent-space model.
  Four different samples per method, each with zoomed-in crops of distinctive regions (sea, sky, texture, rocks).
  }
  \label{fig:lighthouse_comparison}
\end{figure}

In \Cref{fig:lighthouse_comparison} we compare samples generated with WaiT, JiT and RAE for ImageNet 512 resolution class-conditional generations. In this figure, as in \Cref{fig:oi512_comparison,fig:oi512_stainedglass,fig:oi1024_comparison}, the three small images to the right of each sample show, from top to bottom, the downsampled full image, a zoomed crop of a detailed region, and a zoomed crop of a textured region. They illustrate what FID, 5cFID and hFWD are meant to measure (global coherence, local detail and texture fidelity, respectively) and are for illustrative purposes only.

\begin{figure}
  \centering
  \includegraphics[width=\linewidth]{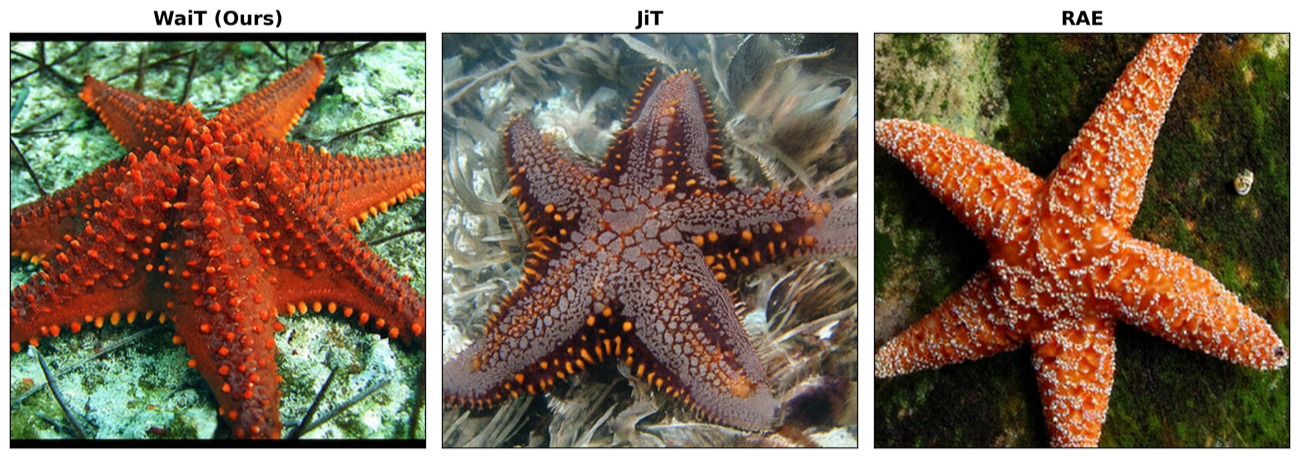}
  \caption{\textbf{Full-resolution starfish samples} on ImageNet 512$\times$512. From left to right: WaiT-H/32 (ours), JiT baseline, RAE.}
  \label{fig:starfish_full_appendix}
\end{figure}

\Cref{fig:starfish_full_appendix} shows additional full-resolution starfish samples.

\subsection{OpenImages}
\label{app:openimages_pareto}
\label{app:openimages512}
\label{app:openimages1024}

\begin{figure}
  \centering
  \includegraphics[width=\linewidth]{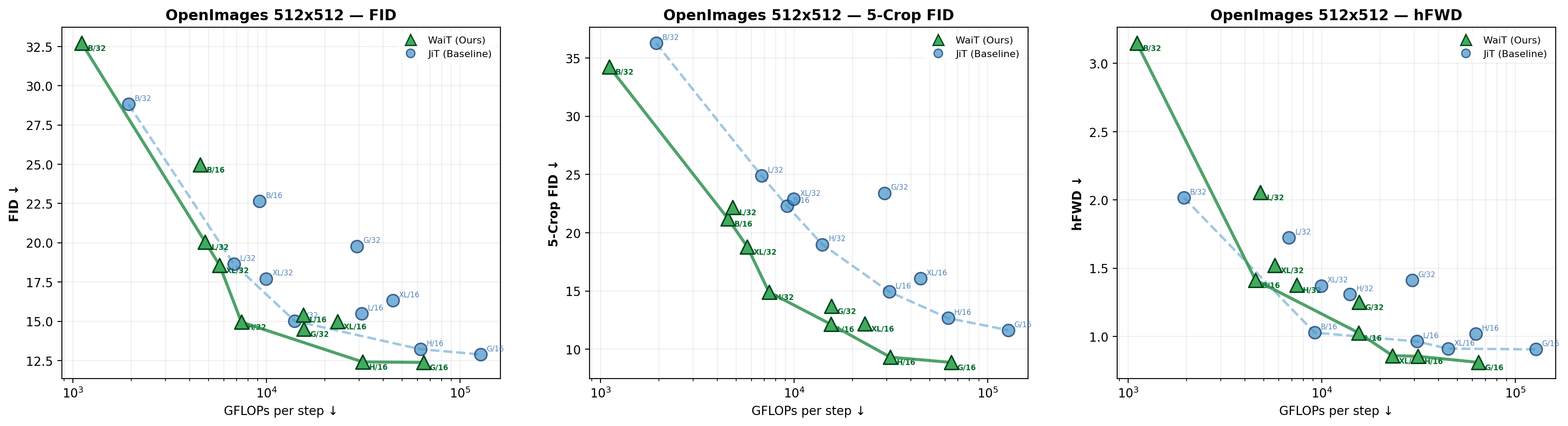}
  \caption{\textbf{OpenImages 512$\times$512 Pareto fronts} (5k train FID): FID, 5-Crop FID, and hFWD. WaiT (green triangles, solid line) dominates the JiT baseline (blue circles, dashed line) on FID and 5-crop FID across model scales, and is comparable to it on hFWD. OpenImages 1024$\times$1024 Pareto fronts are shown in the teaser (\Cref{fig:teaser}).}
  \label{fig:pareto_openimages}
\end{figure}

\Cref{fig:pareto_openimages} shows the Pareto fronts for WaiT and the JiT baseline across all model scales on OpenImages 512  resolution (train FID over 5k samples).
The WaiT Pareto front (green, solid) dominates the one for the JiT baseline (blue, dashed) on FID and 5-crop FID, achieving the same or better performance at 2$\times$ lower compute, and is comparable to it on hFWD.

\begin{figure}
  \centering
  \includegraphics[width=.68\linewidth]{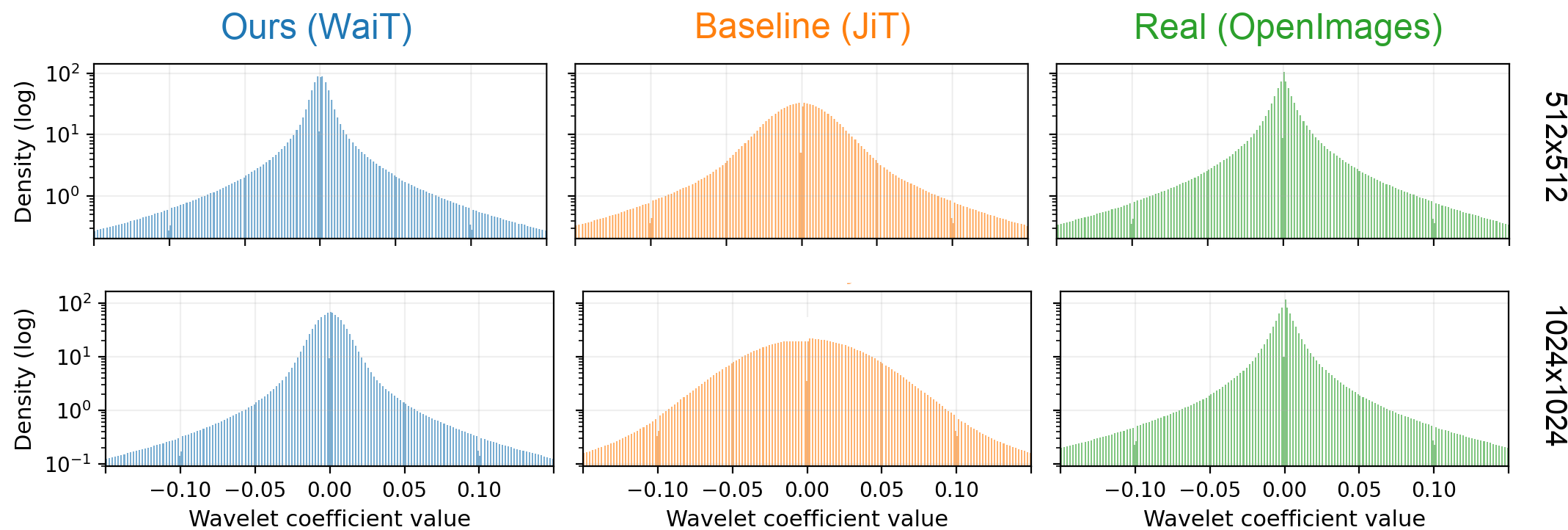}
  \caption{\textbf{High-frequency wavelet coefficient distributions on OpenImages.} At 512$\times$512 our distribution already tracks the real data more closely than JiT, and the gap widens at 1024$\times$1024 where JiT degrades to a markedly flatter, broader distribution -- our advantage grows with resolution.}
  \label{fig:wavelet_hf}
\end{figure}

In \Cref{fig:wavelet_hf} we show wavelet coefficient distributions for OpenImages at 512 and 1024 resolution.

\begin{figure}[htbp]
  \centering
  \includegraphics[width=\linewidth,height=0.85\textheight,keepaspectratio]{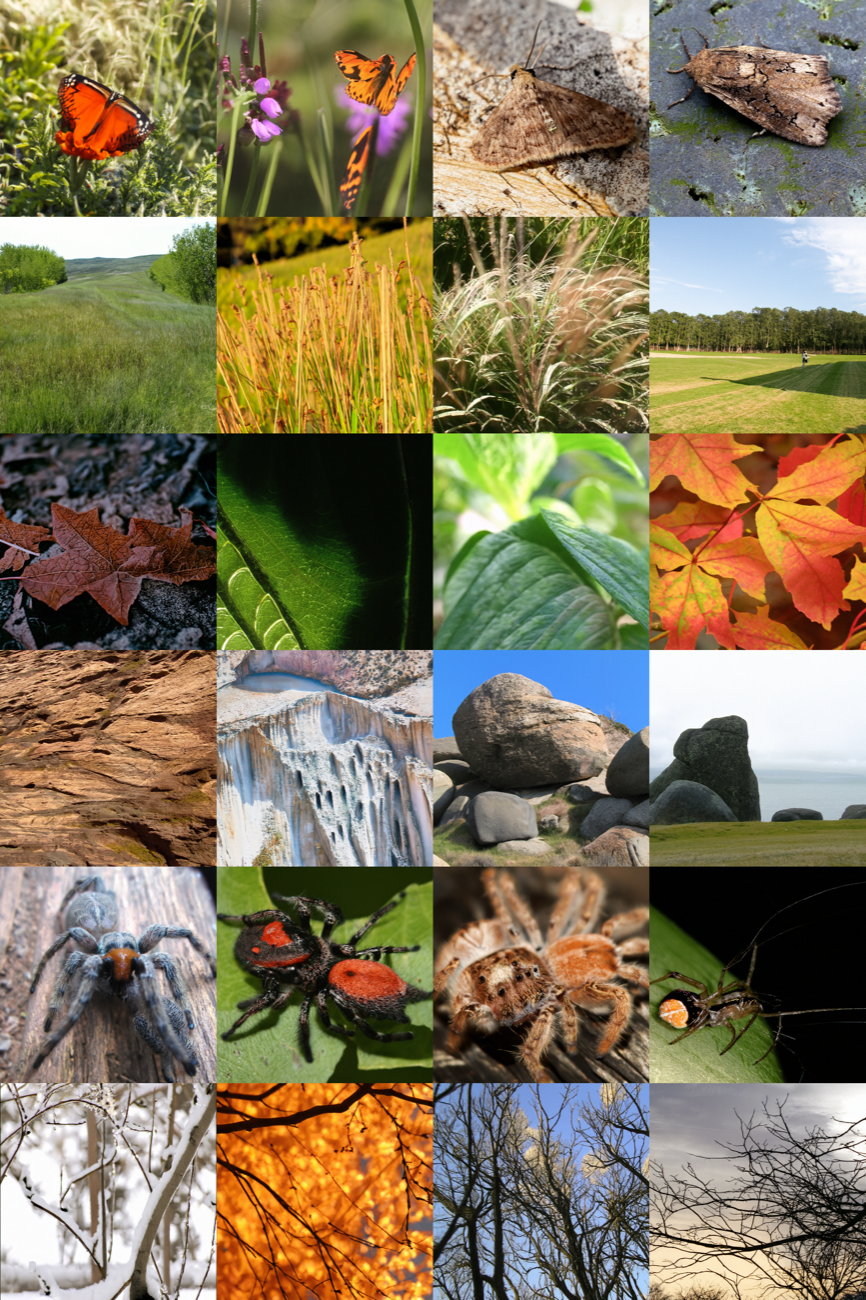}
  \caption{\textbf{Uncurated WaiT-H/32 samples on OpenImages 512$\times$512.} Each row shows four samples from a single class. From top to bottom: \emph{butterfly, grass, leaf, rock, spider, twig.}}
  \label{fig:oi512_generations}
\end{figure}

\begin{figure}
  \centering
  \includegraphics[width=\linewidth]{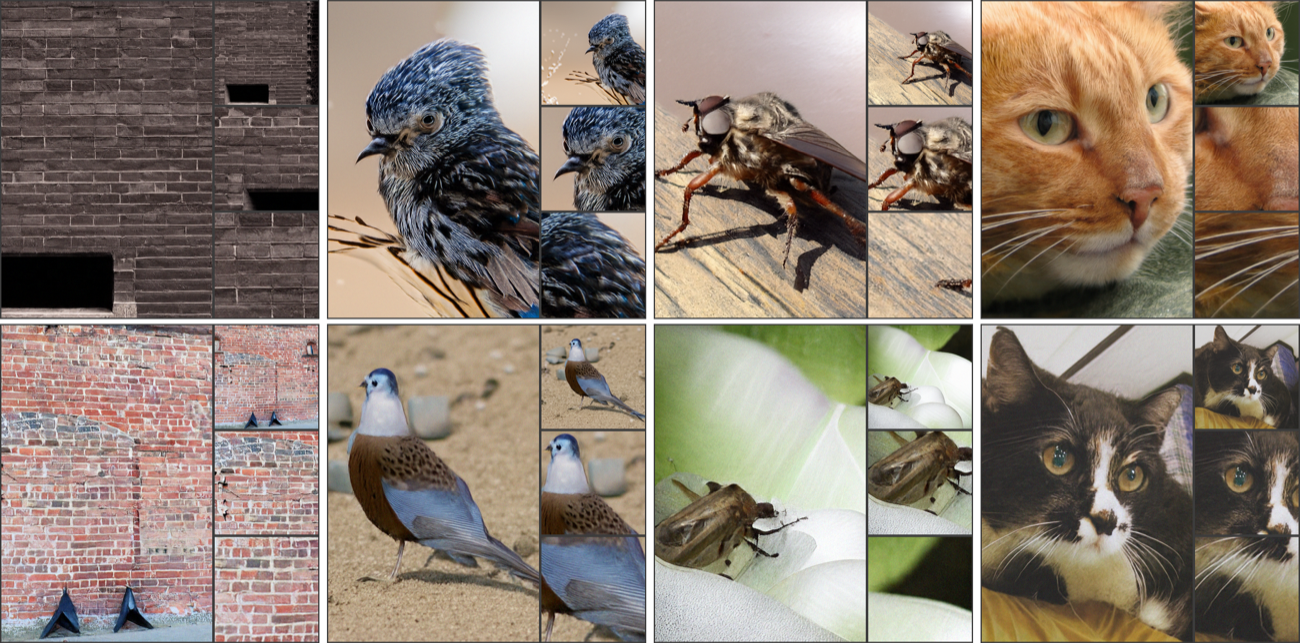}
  \caption{\textbf{Qualitative comparison on OpenImages 512$\times$512.} Each cell shows the generated image with two zoomed-in crops. \textbf{Top:} WaiT-H/32 (ours). \textbf{Bottom:} JiT baseline. Columns: \emph{brick, feather, insect, whiskers.}}
  \label{fig:oi512_comparison}
\end{figure}

\begin{figure}
  \centering
  \includegraphics[width=\linewidth]{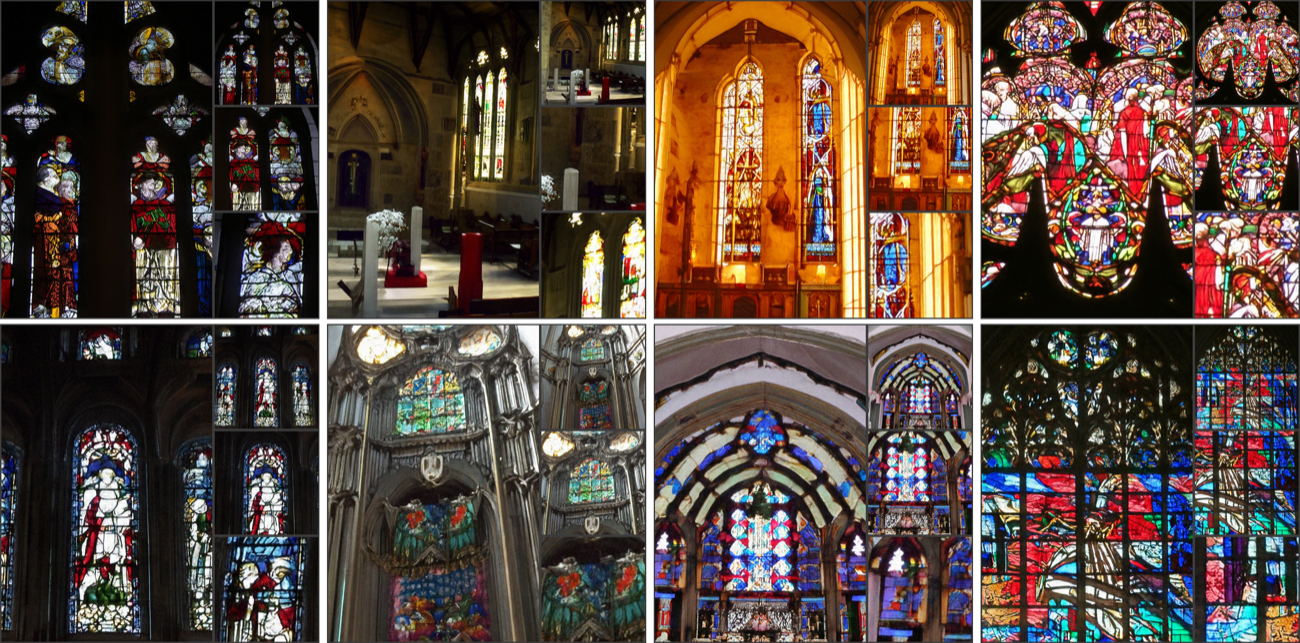}
  \caption{\textbf{Single-class comparison on OpenImages 512$\times$512.} Four samples from the class \emph{stained glass} for WaiT-H/32 (top) and  JiT baseline (bottom).
  }
  \label{fig:oi512_stainedglass}
\end{figure}

In \Cref{fig:oi512_generations} we provide samples from WaiT for OpenImages at 512 resolution.
In \Cref{fig:oi512_comparison} we show more WaiT samples and compare to JiT and RAE.
\Cref{fig:oi512_stainedglass} compares WaiT and JiT using four samples for a single class.

\begin{figure}[htbp]
  \centering
  \includegraphics[width=0.9\linewidth]{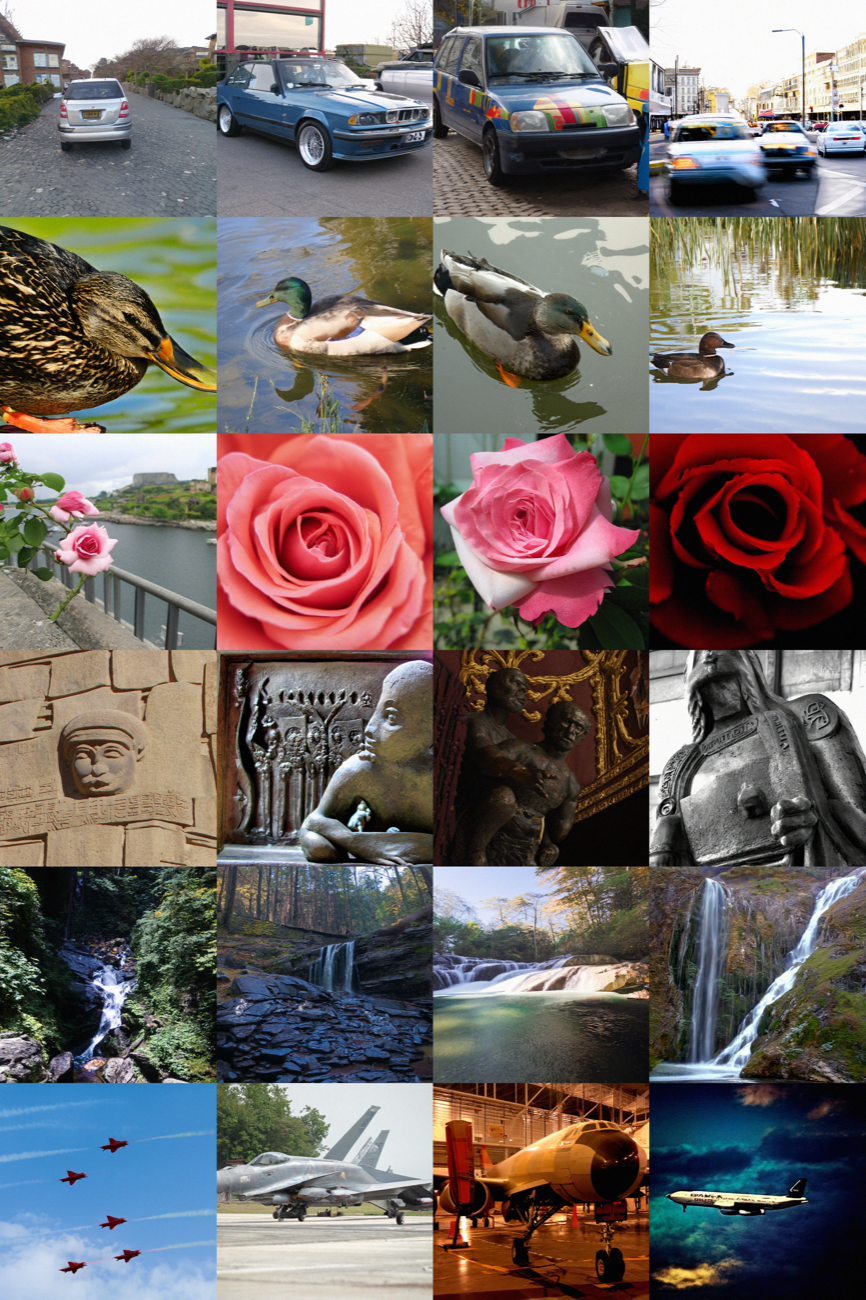}
  \caption{\textbf{Uncurated samples from WaiT-H/64 for OpenImages 1024$\times$1024.} Each row shows four samples from a single class, from top to bottom: \emph{car, duck, rose, sculpture, waterfall, airplane.}
  }
  \label{fig:oi1024_generations}
\end{figure}

\begin{figure}[htbp]
  \centering
  \includegraphics[width=\linewidth]{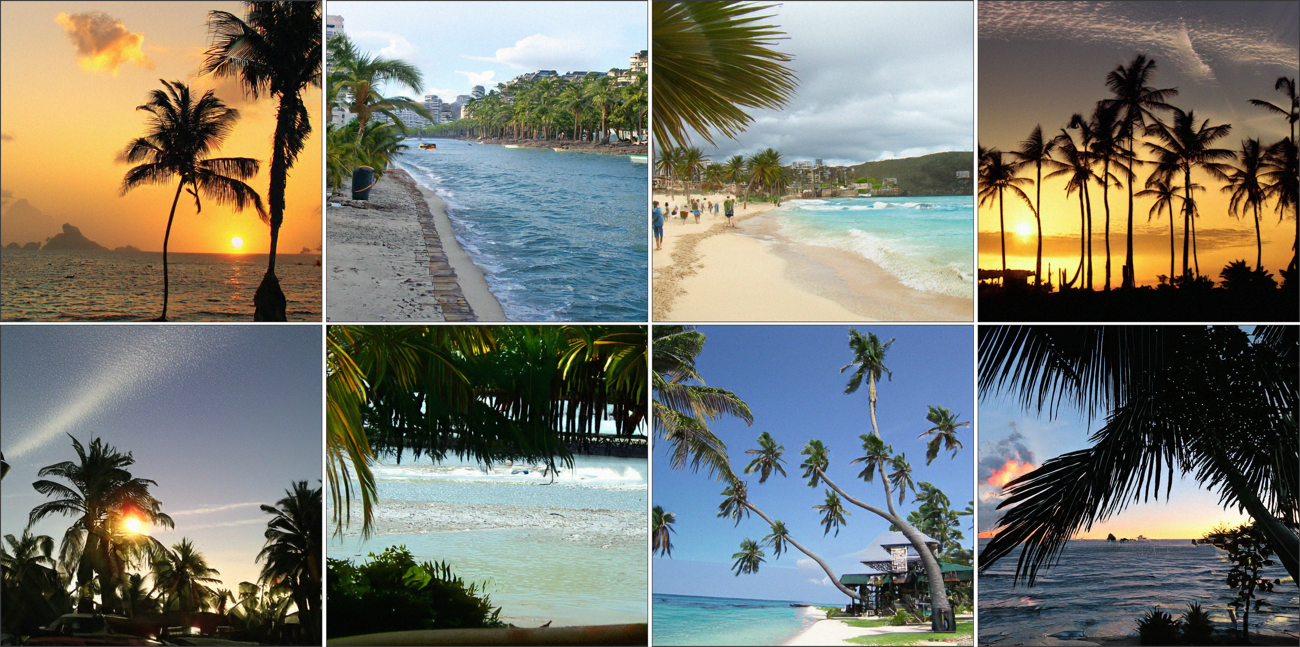}
  \caption{\textbf{Single-class comparison  on OpenImages 1024$\times$1024.} Samples for the class \emph{palm tree} for WaiT-H/64 (ours, top), and  JiT-H/64 baseline (bottom).}
  \label{fig:oi1024_palmtree}
\end{figure}

\begin{figure}[htbp]
  \centering
  \includegraphics[width=\linewidth]{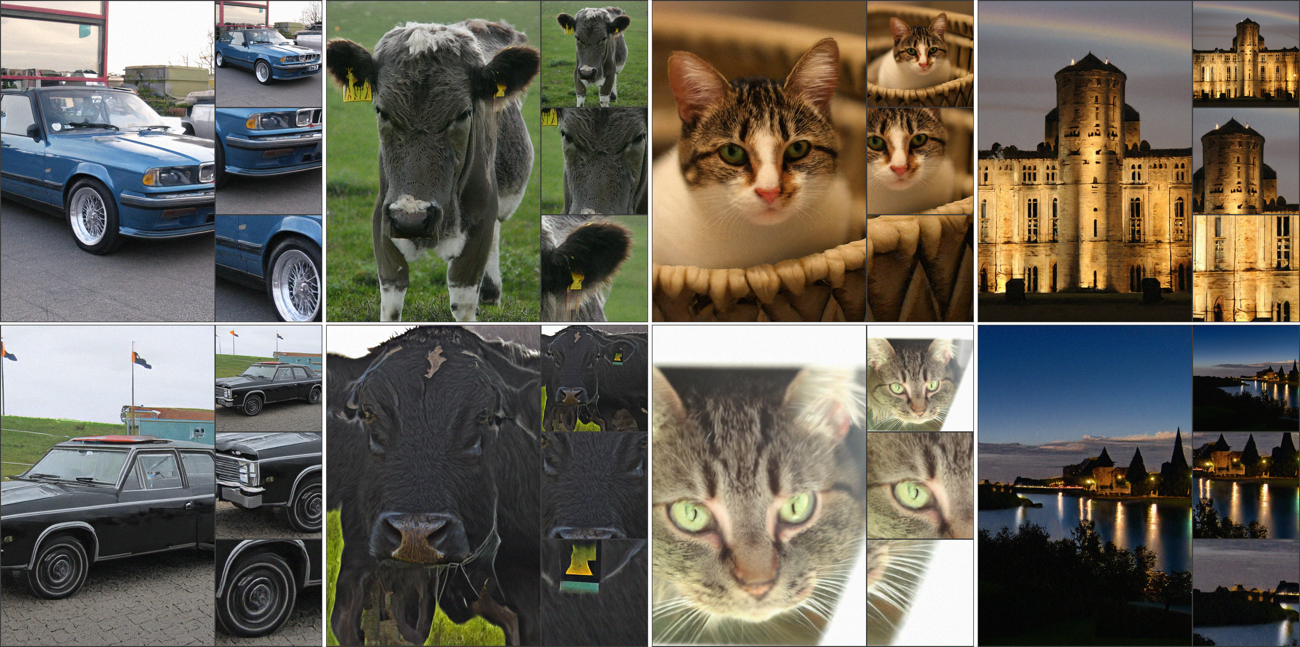}
  \caption{\textbf{Qualitative comparison on OpenImages 1024$\times$1024.} Each cell shows the generated image with two zoomed-in crops. \textbf{Top:} WaiT-H/64 (ours). \textbf{Bottom:} JiT-H/64 baseline. Columns: \emph{car, cattle, cat, castle.} WaiT produces sharper textures (e.g., cattle fur, cat whiskers) alongside better global structure.}
  \label{fig:oi1024_comparison}
\end{figure}

In \Cref{fig:oi1024_generations} we provide samples from WaiT for OpenImages at 1024 resolution.
\Cref{fig:oi1024_palmtree} compares WaiT and JiT using four samples for a single class, and \Cref{fig:oi1024_comparison} across four classes.

\subsection{Text-to-image samples (extended captions)}
\label{app:t2i_captions}
\ifmeta\else\tiifigure\fi

This subsection spells out the per-column / per-row details of the text-to-image figures and provides the additional uncurated samples.

\paragraph{\Cref{fig:t2i_combined} (\Cref{app:t2i}).}
Both rows show 1024$\times$1024 generations from H/32 models on a held-out evaluation set; the \emph{top row} is the JiT-H/32 baseline and the \emph{bottom row} is our WaiT-H/32. The eight columns correspond to two encoder pipelines:
\textbf{Columns 1--4} use the SigLIP\,+\,Qwen3-VL\,+\,T5-v1.1-XXL pipeline (50 training epochs).
\textbf{Columns 5--8} use the MetaCLIP\,+\,Llama~3.2~Vision\,+\,Llama~3.1 pipeline (50 training epochs).
Within each pipeline the four columns show the same set of prompts for the baseline and our method, so column-wise comparisons are like-for-like.
The prompts (left to right) are:
\begin{itemize}
  \item \textbf{Col 1} (SigLIP): \emph{``a hummingbird hovering in front of a fuchsia flower, wings frozen mid-beat''}.
  \item \textbf{Col 2} (SigLIP): \emph{``a Venetian gondolier in a striped shirt rowing through a narrow canal at sunset''}.
  \item \textbf{Col 3} (SigLIP): \emph{``the interior of a grand Moorish palace with intricate geometric tile work''}.
  \item \textbf{Col 4} (SigLIP): \emph{``a stack of fluffy pancakes with maple syrup dripping down the sides''}.
  \item \textbf{Col 5} (MetaCLIP): \emph{``A solitary lighthouse on a rocky cliff during a thunderstorm, waves crashing below, lightning in the background''}.
  \item \textbf{Col 6} (MetaCLIP): \emph{``A detailed macro photograph of a peacock feather with iridescent blue and green patterns''}.
  \item \textbf{Col 7} (MetaCLIP): \emph{``A hand-painted ceramic bowl filled with fresh berries, morning dew still visible on them''}.
  \item \textbf{Col 8} (MetaCLIP): \emph{``A street in Havana with colorful vintage cars, pastel colonial buildings, and palm trees''}.
\end{itemize}

\begin{figure}[htbp]
  \centering
  \includegraphics[width=\linewidth]{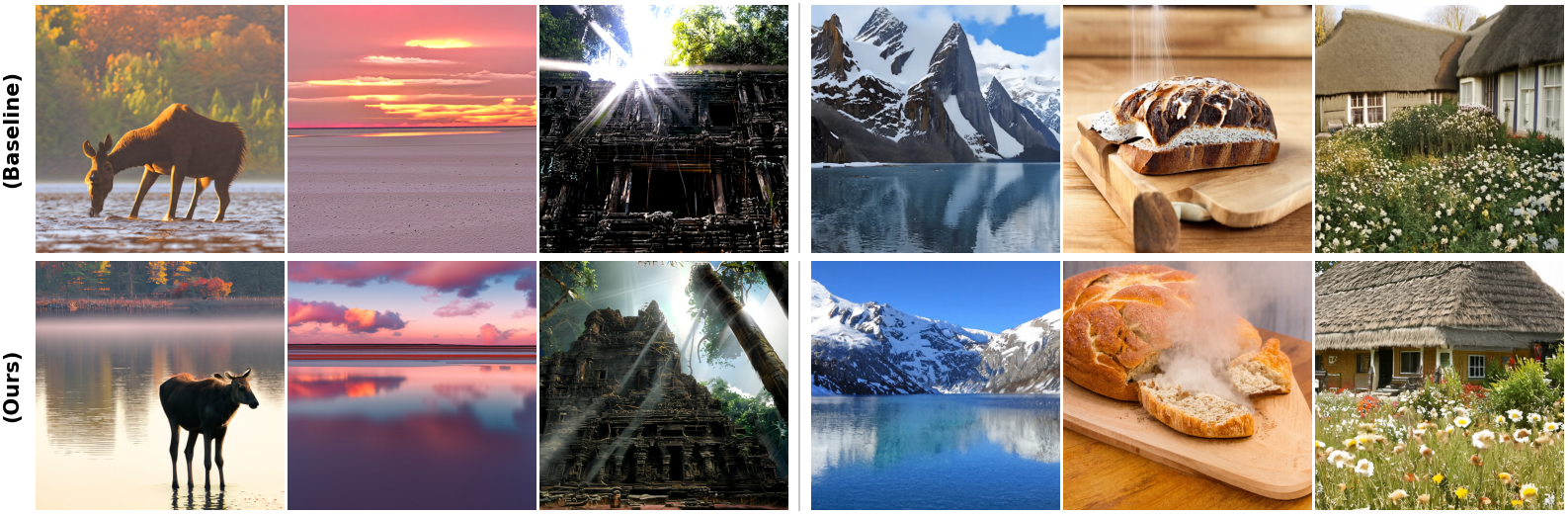}
  \caption{Additional text-to-image samples at 1024$\times$1024. See \Cref{app:t2i_captions} above for the per-column / per-row description.}
  \label{fig:t2i_appendix}
\end{figure}

\paragraph{\Cref{fig:t2i_appendix} (additional samples).}
Same setup as \Cref{fig:t2i_combined}, on the remaining columns not shown in \Cref{fig:t2i_combined}.
\textbf{Columns 1--3}: SigLIP\,+\,Qwen3-VL\,+\,T5-v1.1-XXL.
\textbf{Columns 4--6}: MetaCLIP\,+\,Llama~3.2~Vision\,+\,Llama~3.1.
\textbf{Top row:} JiT-H/32 (baseline). \textbf{Bottom row:} WaiT-H/32 (ours).
The prompts (left to right) are:
\begin{itemize}
  \item \textbf{Col 1} (SigLIP): \emph{``a moose standing in a misty morning lake with autumn foliage in the background''}.
  \item \textbf{Col 2} (SigLIP): \emph{``an endless glowing salt flat at dusk reflecting pink and orange clouds''}.
  \item \textbf{Col 3} (SigLIP): \emph{``an ancient ruined temple completely overtaken by jungle roots''}.
  \item \textbf{Col 4} (MetaCLIP): \emph{``A crystal clear mountain lake reflecting snow-capped peaks and a cloudless blue sky''}.
  \item \textbf{Col 5} (MetaCLIP): \emph{``A freshly baked sourdough loaf on a wooden cutting board, steam rising from a torn piece''}.
  \item \textbf{Col 6} (MetaCLIP): \emph{``A quaint English cottage with a thatched roof surrounded by a wildflower garden in full bloom''}.
\end{itemize}

\subsection{Video generation samples}
\label{app:video_qualitative}
\ifmeta\else
\begin{table}[h]
\kineticstable
\end{table}
\fi

\begin{figure}[htbp]
  \centering
  \includegraphics[width=\linewidth]{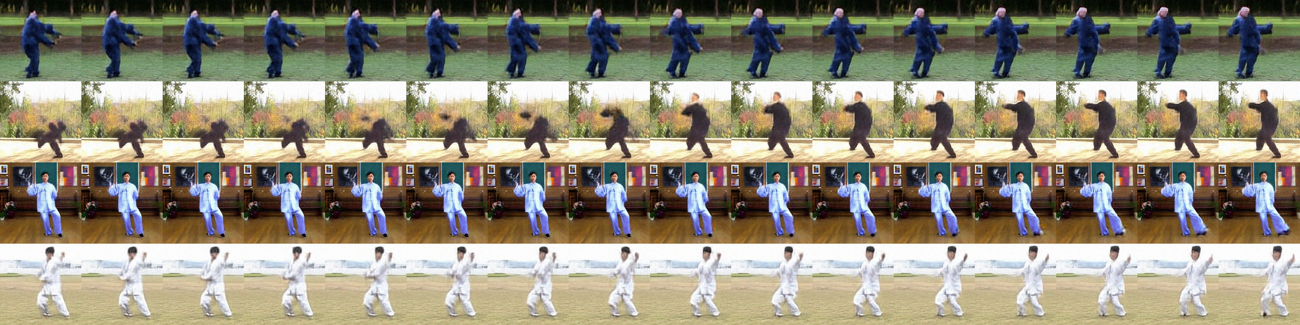}
  \caption{\textbf{Unconditional Taichi-HD samples} from our wavelet-aware image transformer WaiT-B/8 model (128x128, 16 frames).
  }
  \label{fig:taichi_samples}
\end{figure}

\begin{figure}[htbp]
  \centering
  \includegraphics[width=\linewidth]{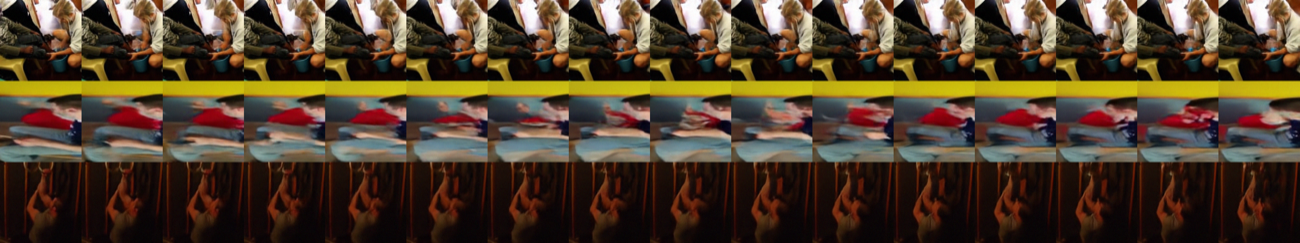}
  \caption{\textbf{Samples of WaiT-B/8 trained on  Kinetics-600.}
  Conditioned on first 5 frames, generates the remaining 11.
  }
  \label{fig:kinetics_samples}
\end{figure}

In \Cref{fig:taichi_samples} we provide unconditional samples from WaiT-B/8 trained on Taichi-HD.
In \Cref{fig:kinetics_samples} we provide samples for Kinetics-600.

\section{Dataset Construction Details}
\label{sec:dataset_construction}

This section contains the dataset construction details needed to reproduce the OpenImages-512 ($512{\times}512$) and OpenImages-1024 ($1024{\times}1024$) datasets, as well as the corpus used in our text-to-image experiments, which is a mixture of several large public datasets. All components from which we construct these datasets are publicly available. We first describe the OpenImages datasets (\Cref{sec:source_data,sec:oi512}) and then the text-to-image corpus (\Cref{app:t2i_data}).

\subsection{Source Data}
\label{sec:source_data}

{We use OpenImages V6~\cite{kuznetsova2020openimages}, hosted by Google. For training we use the train images obtained through the download page\ifmeta\ {\small\url{https://storage.googleapis.com/openimages/web/download_v6.html}}\fi.

Our goal is to construct a dataset that resembles ImageNet-1k as closely as possible, i.e.\ roughly balanced over $1{,}000$ classes with around $1$M images, but whose images are always at full resolution and never need upsampling: images larger than the target resolution are resized and center-cropped.

\paragraph{Labels.} OpenImages does not contain enough human-verified labels to build such a dataset, so we combine the human-verified labels with the machine-generated ones.
\begin{itemize}[leftmargin=1.4em,topsep=2pt,itemsep=1pt]
\item \emph{Human-verified labels:} \texttt{oidv6-train-annotations-human-imagelabels.csv}, of which we only use the labels for which a human annotator confirmed that the class is present in the image (attribute $\texttt{Confidence}{=}1$).\ifmeta\\
{\small\url{https://storage.googleapis.com/openimages/v6/oidv6-train-annotations-human-imagelabels.csv}}\fi
\item \emph{Machine-generated labels:} \texttt{train-annotations-machine-imagelabels.csv}, found on the download page under ``Subset with Image-Level Labels'' $\rightarrow$ Machine-generated labels $\rightarrow$ Train. Each label comes with a confidence score in $[0,1]$.
\item \emph{Class descriptions:} found on the download page under Metadata $\rightarrow$ Class Names\ifmeta, or at {\small\url{https://storage.googleapis.com/openimages/v5/class-descriptions-boxable.csv}}\fi.
\end{itemize}}

\subsection{OpenImages-512 and OpenImages-1024 Construction}
\label{sec:oi512}

We build both datasets with exactly the same process; only the target resolution $R$ ($512$ or $1024$) and two parameters differ, as stated below.

\paragraph{Stage 1: Manifest building.} We first discard every source image whose shorter side is below $R$ pixels, so that we never need to upsample: an upsampled image has no real content in the upper part of its frequency spectrum, which is exactly what we want the model to learn. Now, we take all classes, rank them by the number of human-verified labels and keep the top $1{,}000$. For every image that has multiple labels, we assign it the rarest class, i.e.\ the class with the fewest images; if multiple classes share the same count, we just take them in the order in which their labels appear in the original source CSV. We cap every class at $1{,}280$ images for OpenImages-512, to match ImageNet's ${\sim}1{,}280$ images per class, and at $1{,}500$ images for OpenImages-1024. For classes below the cap, we add machine-labeled images, taking the most confident ones first and only those with confidence at least $0.90$ for OpenImages-512 and $0.70$ for OpenImages-1024, in order to attempt avoiding incorrect labels. The lower threshold at $1024$ compensates for the smaller pool of high-resolution images. We then drop all classes that still have fewer than $500$ images.

\paragraph{Stage 2: Image processing.} About $1\%$ of the OpenImages source files are grayscale or CMYK, so we convert all images to RGB. We then resize each image so that its shorter side is $R$ pixels and center-crop it to $R{\times}R$. The resize is needed because most source images are much larger than $R$ pixels (median shorter side ${\sim}1{,}800$ pixels); a center crop alone would keep only a small central patch of the image. We use Lanczos resampling, which preserves fine detail better than bilinear or bicubic. Each image is saved as a JPEG with quality $100$ (the lowest compression level), without chroma subsampling (the color channels are stored at full resolution), and without an embedded color profile (so that all images are decoded the same way).

\subsection{Text-to-image dataset construction}
\label{app:t2i_data}
We build the $1024\times1024$ text-to-image corpus entirely from publicly available sources using publicly available models; the corpus is fully reconstructable by anyone from scratch. We start from three public sources: SA-1B~\citep{kirillov2023sam}, the DataComp~\citep{gadre2024datacomp}, and OpenImages~\citep{kuznetsova2020openimages}. The pipeline has three stages. We describe the SigLIP\,+\,Qwen3-VL\,+\,T5 configuration; the second (MetaCLIP\,+\,Llama) configuration is identical with the corresponding models substituted.

\medskip\noindent\textbf{Stage 1 (resolution filter).} We scan all sources and keep only images whose shorter side is at least $1024$ pixels, parsing image headers without decoding pixels. All images are resized so that their shorter side is $1024$ pixels and center-cropped to $1024{\times}1024$; images are then stored as JPEG with quality $85$ and 4:2:0 chroma subsampling.

\medskip\noindent\textbf{Stage 2 (aesthetic and watermark scoring).} Each remaining image is scored zero-shot with a public vision--language model (SigLIP~2, \texttt{google/siglip2-large-patch16-384}; MetaCLIP in the second pipeline). For each of the two concepts we use a small set of positive and negative text prompts, take the mean image--text cosine similarity per side, and apply a softmax to obtain a probability in $[0,1]$:
\begin{itemize}[leftmargin=1.4em,topsep=2pt,itemsep=1pt]
\item \emph{Aesthetic}---positive: ``a beautiful high quality photograph''; ``a professional photograph with great composition''; ``an aesthetically pleasing image with good lighting''. Negative: ``a low quality blurry photograph''; ``an ugly poorly composed image''; ``a low resolution amateur snapshot''.
\item \emph{Watermark}---positive: ``an image with a watermark''; ``a photo with text overlay''; ``an image with a logo watermark''; ``a stock photo with watermark text''. Negative: ``a clean photograph without watermarks''; ``a natural photo without text overlay''.
\end{itemize}
We then retain images with $\mathrm{aesthetic\_prob} > 0.05$ and $\mathrm{watermark\_prob} < 0.80$, leaving the curated corpus of ${\sim}40$M images reported in \Cref{app:t2i}.

\medskip\noindent\textbf{Stage 3 (captioning).} Each retained image is resized to $448\times448$ and captioned by a public VLM (Qwen3-VL-30B-A3B-Instruct; Llama-3.2-Vision in the second pipeline) with greedy decoding ($\texttt{max\_new\_tokens}{=}160$, target ${\sim}60$ words) and the prompt: \emph{``Describe this image in a single vivid paragraph of approximately 60 words. Focus on the main subject, its attributes, the setting or background, colors, lighting, composition, and overall mood. Do not use bullet points. Do not add preamble like `This image shows'; go straight to the description.''}

\medskip\noindent\textbf{Text conditioning.} At training time captions are encoded on the fly by a frozen text encoder (T5-v1.1-XXL; Llama-3.1 in the second pipeline) with a maximum length of $128$ tokens, and the model is conditioned on the resulting embeddings via cross-attention within each transformer block. A cached null-caption embedding is used for classifier-free guidance with text-dropout probability $0.1$.

\stopcontents[appendices]

\end{document}